\documentclass[11pt]{article}

\PassOptionsToPackage{table}{xcolor}
\usepackage[
  shownumpages,            %
  citingstyle=numbers,     %
]{brainlab}

\usepackage{shortcuts}
\usepackage{microtype}
\usepackage{booktabs}
\usepackage{multirow}
\usepackage{graphicx}
\usepackage{float}


\newcommand{\new}[1]{\textcolor{blue}{#1}}
\renewcommand{\new}[1]{#1}

\setbrainmeta{
  title={WRDO: Weighted Refusal Direction Optimization in Aligned Language Models},
  authors={
    Grigoriy Evseev\textsuperscript{1,2},
    Vladimir Solodkin\textsuperscript{1,2,3},
    Dmitry Bylinkin\textsuperscript{1,2},
    Daniil Medyakov\textsuperscript{1,2},
    Vladimir Aletov\textsuperscript{1,2},
    Aleksandr Beznosikov\textsuperscript{1,2,4}
  },
  affiliations={
    \textsuperscript{1}Moscow Independent Research Institute of Artificial Intelligence \\
    \textsuperscript{2}Basic Research of Artificial Intelligence Laboratory (BRAIn Lab) \\
    \textsuperscript{3}T-Technologies \\
    \textsuperscript{4}Innopolis University \\
  },
  abstract={
    Aligned large language models are trained to comply with benign requests while refusing harmful ones. However, recent work shows that this refusal behavior is fragile and can often be disrupted by simple activation- or weight-space interventions. In particular, refusal has been shown to be largely mediated by a single residual-stream direction, typically computed as a difference-of-means vector between harmful and harmless prompts. However, this standard construction aggregates heterogeneous harmful prompts into one uniform average, potentially conflating distinct refusal-related phenomena. In this work, we study whether refusal interventions can be strengthened by learning how to combine prompt-specific refusal directions. We propose Weighted Refusal Direction Optimization (WRDO), a procedure that constructs a composite refusal direction from a bank of per-prompt directions using learned aggregation weights. This approach allows different harmful prompts to contribute unequally to the final intervention direction. We evaluate WRDO against prior single-direction methods on three open-weight language models using three guard judges and HarmBench as an external evaluator. We further measure the effect of the intervention on general model capability through standard utility benchmarks. Finally, we investigate a defense-oriented setting in which the same optimization machinery is used in reverse to restore refusal behavior in an abliterated model checkpoint. Our results show that weighted aggregation provides a more effective and flexible mechanism for manipulating refusal representations. Code is available at \url{https://github.com/brain-lab-research/WRDO}
  },
}

\begin{document}
\begin{mainpart}

\section{Introduction}
\label{sec:intro}

Large language models deployed in user-facing settings are aligned to follow benign instructions and to refuse harmful ones, with this behavior shaped by a combination of supervised fine-tuning, instruction tuning and preference optimization \citep{ouyang2022training,bai2022training}. The resulting refusal mechanism is, in practice, brittle under both prompt-level adversarial attacks \citep{zou2023gcg,wei2023jailbroken} and direct interventions on the model itself, which has motivated a mechanistic line of work that asks how refusal is represented internally and which components of the model implement it. A central finding of this line is that refusal is, to a first approximation, mediated by a single direction in the residual stream of the transformer \citep{arditi2024refusal}: ablating one difference-of-means vector, computed between activations on harmful and harmless prompts at a chosen layer, is sufficient to disable refusal in a broad set of open-weight aligned models. The same work establishes an equivalence between activation ablation and weight-space orthogonalization of the attention output and feed-forward down projections, which makes the intervention a one-shot edit of the model with no inference-time overhead, and public toolkits such as Heretic \citep{heretic} operationalize this recipe end-to-end.

The standard procedure averages every harmful prompt into a single axis. A disinformation prompt, a malware prompt and a privacy-violation prompt all enter the same mean, and the resulting direction is diffuse in the sense that it conflates phenomena that need not be aligned in the residual stream. A natural alternative is to retain a per-prompt direction $r_i$ for each training prompt and to combine them with a coefficient vector $w$. The simplest combination is the uniform average, $r = \mathrm{normalize}(\sum_i r_i / N)$, but there is no a priori reason for every prompt to contribute equally. \new{Performance-aware weighted aggregation is known to beat uniform averaging whenever the aggregated components differ in quality, from ensembling ranked classifiers \citep{ahsen2019weighted} to federated averaging \citep{shaikh2026fedwapr} and adapter selection \citep{veprikov2026weightlora}, and per-prompt refusal directions are heterogeneous in exactly that way.} In this work, we research the question of whether computing refusal direction using the aggregation weights yields a stronger intervention than the single-direction recipe.

\paragraph{Our Contributions:}
\begin{itemize}
    \item We propose Weighted Refusal Direction Optimization (WRDO), a procedure that obtains a refusal direction by learning aggregation weights $w$ over a bank of per-prompt directions, together with two concrete instantiations of WRDO based on separable CMA-ES and a proposed GRPO variant with importance sampling (GRPO-IS).
    \item We show empirically that the proposed WRDO instantiations outperform all considered prior single-direction baselines across three open-weight models and three guard judges, and that this superiority transfers to HarmBench ruling out judge-specific overfitting.
    \item We obtain that WRD-GRPO-IS preserves the general capability of the intervened model substantially better than every other considered method on standard utility benchmarks.
    \item We additionally study a defense setup that reuses the same machinery in reverse to restore refusal on an abliterated checkpoint, illustrating an alternative use of the proposed approach.
\end{itemize}

The remainder of the paper is organized as follows. Section~\ref{sec:formulation} formalizes the WRDO problem. Section~\ref{sec:methods} presents the two WRDO instantiations introduced in this work, namely an adaptation of separable CMA-ES and a proposed GRPO variant with importance sampling, together with the three single-direction baselines against which they are compared. Section~\ref{sec:design} specifies the experimental setup, and Sections~\ref{sec:results}--\ref{sec:conclusion} report the empirical results, ablations and discussion.

\section{Problem Formulation}
\label{sec:formulation}

An aligned language model is denoted as $f_\theta$, a harmful training set of prompts $\mathcal{D}_{\text{train}}^{\text{harm}} = \{x_i\}_{i=1}^N$, a held-out harmful validation set $\mathcal{D}_{\text{val}}^{\text{harm}}$ and a harmless baseline corpus $\mathcal{D}_{\text{benign}}$. The hidden dimensionality of the residual stream is $d$, and $L$ denotes the number of transformer blocks.

\subsection{Candidate bank construction}
\label{sec:bank}

Let $h^{(l)}(x) \in \mathbb{R}^d$ denote the residual-stream activation of $f_\theta$ on prompt $x$ at layer $l$, evaluated at the last token position. Activations are extracted at every transformer block, including the embedding output, so that each prompt is summarized by a tensor in $\mathbb{R}^{(L+1) \times d}$. The mean benign activation $\mu_{\text{benign}}^{(l)} = \mathbb{E}_{x \sim \mathcal{D}_{\text{benign}}}[h^{(l)}(x)]$ is computed at every layer.

For each training prompt $x_i$ we define a per-prompt layerwise direction by taking the difference $h^{(l)}(x_i) - \mu_{\text{benign}}^{(l)}$ and applying $L_2$ normalization at every layer independently,
\begin{equation}
r_i^{(l)} = \frac{h^{(l)}(x_i) - \mu_{\text{benign}}^{(l)}}{\bigl\|h^{(l)}(x_i) - \mu_{\text{benign}}^{(l)}\bigr\|_2},
\label{eq:per-prompt-dir}
\end{equation}
and stack the per-layer vectors into a layerwise tensor $r_i \in \mathbb{R}^{(L+1) \times d}$. The candidate bank $\mathcal{R} = \{r_i\}_{i=1}^N$ contains one such tensor per harmful training prompt, with $N = |\mathcal{D}_{\text{train}}^{\text{harm}}|$. The bank is computed once per model and is shared across all optimization methods.

\subsection{Weighted direction}
\label{sec:wdir}

A coefficient vector $w \in \mathbb{R}^N$ aggregates the bank of Section~\ref{sec:bank} by a scalar weighting that is shared across layers and across hidden coordinates, followed by a per-layer $L_2$ renormalization. Concretely, the composite direction at layer $l$ is
\begin{equation}
r^{(l)}(w) = \frac{\sum_{i=1}^{N} w_i\, r_i^{(l)}}{\bigl\|\sum_{i=1}^{N} w_i\, r_i^{(l)}\bigr\|_2}.
\label{eq:rw}
\end{equation}
The per-layer normalization enforces $\|r^{(l)}(w)\|_2 = 1$ at every layer and removes the overall positive scale of $w$, so that only the orientation of the composite at each layer is varied. Intervention intensity is controlled separately through the layerwise schedule described in Section~\ref{sec:operator}. The single-direction baseline of \citep{arditi2024refusal} is recovered, in this notation, by replacing the per-prompt $r_i$ by a single difference-of-means tensor (computed between the mean harmful and mean benign activations at every layer) and setting $w_1 = 1$ with $N=1$.

\subsection{Intervention operator}
\label{sec:operator}

The primary intervention is weight-space orthogonalization. We adopt the intervention operator of the Heretic toolkit \citep{heretic}. Given the layerwise composite direction $r^{(l)}(w)$ of Equation~\ref{eq:rw}, every attention output projection $W_O^{(l)}$ and every feed-forward down-projection $W_{\text{down}}^{(l)}$ at block $l$ is replaced by
\begin{equation}
W^{(l)\prime} = W^{(l)} - \alpha^{(l)}\, r^{(l)}(w)\, r^{(l)}(w)^\top W^{(l)},
\label{eq:operator}
\end{equation}
where $\alpha^{(l)}$ is the fixed layerwise intensity schedule. The full form of $\alpha^{(l)}$, together with the values of the auxiliary parameters that we hold fixed throughout the paper, is given in Appendix~\ref{app:schedule}. The only departure from the Heretic recipe is the choice of $r^{(l)}(w)$: Heretic plugs in the single direction of \citep{arditi2024refusal}, whereas WRDO substitutes the per-prompt composite of Equation~\ref{eq:rw}.

The composition of $f_\theta$ with the operator of Equation~\ref{eq:operator}, evaluated at the layerwise direction $r(w) = \{r^{(l)}(w)\}_l$ and a given $\alpha$, is denoted $T_{r(w)}(\theta)$. We write $\pi_{T_{r(w)}(\theta)}(\cdot \mid x)$ for the conditional distribution over generations produced by the intervened model under greedy decoding.

\subsection{Harmfulness metric}
\label{sec:objective}

For each prompt $x$ and response $y$, every guard model used in this paper outputs a binary safe/unsafe decision $J(x, y) \in \{0, 1\}$, with $J = 1$ denoting an unsafe response. The attack success rate (ASR) of the intervened model is the average of this per-prompt binary signal over the validation distribution,
\begin{equation}
\mathrm{ASR}_J(w) = \mathbb{E}_{x \sim \mathcal{D}_{\text{val}}^{\text{harm}}}\!\bigl[J(x, \pi_{T_{r(w)}(\theta)}(\cdot|x))\bigr].
\label{eq:asr}
\end{equation}
In practice, $\mathrm{ASR}_J$ is estimated on a validation batch as the empirical mean of the per-prompt decisions.

The optimization objective is the negative attack success rate,
\begin{equation}
\mathcal{L}_J(w) = -\mathrm{ASR}_J(w),
\label{eq:loss}
\end{equation}
which is shared by every WRDO method. Some optimizers may augment this objective with method-specific regularizers when needed (see WRD-GRPO-IS in Section~\ref{sec:grpo} for a KL-regularized variant).
The attack setup minimizes $\mathcal{L}_J$,
\begin{equation}
w^\star = \arg\min_{w \in \mathcal{W}} \mathcal{L}_J(w),
\label{eq:attack}
\end{equation}
where the feasible set $\mathcal{W} \subset \mathbb{R}^N$ is a per-coordinate box on $w$, with bounds and initialization detailed in Appendix~\ref{app:constraints}.

\section{Methods}
\label{sec:methods}

\subsection{Baselines}
\label{sec:baselines}

\paragraph{Basic RD.}
We refer to the single-direction refusal-ablation procedure of \citep{arditi2024refusal} as Basic RD throughout the paper. A single layerwise direction is computed as the per-layer $L_2$-normalized difference of means
\begin{equation}
r_{\text{Basic}}^{(l)} = \frac{\mu_{\text{harm}}^{(l)} - \mu_{\text{benign}}^{(l)}}{\|\mu_{\text{harm}}^{(l)} - \mu_{\text{benign}}^{(l)}\|_2}
\end{equation}
on the training split, where $\mu_{\text{harm}}^{(l)}$ is the mean activation of the harmful training prompts at layer $l$ and $\mu_{\text{benign}}^{(l)}$ is the mean activation of the benign baseline corpus. The resulting layerwise direction is applied through the operator of Equation~\ref{eq:operator}. There is no candidate bank and no optimization. This is the reference against which any benefit of per-prompt aggregation must be measured.

\paragraph{Heretic.}
We additionally include the public Heretic toolkit \citep{heretic} as a further baseline. Heretic applies the same weight-space orthogonalization of Equation~\ref{eq:operator} to the single Basic RD direction, but tunes the layerwise intensity schedule with its own internal Bayesian search rather than using the fixed schedule of Appendix~\ref{app:schedule}. There is no per-prompt candidate bank and no learned $w$; the baseline isolates the contribution of schedule tuning alone, independent of the per-prompt aggregation introduced by WRDO. We run the released pipeline out of the box, with default hyperparameters.

\paragraph{RDO.}
We further include the gradient-based refusal-direction baseline of \citep{wollschlager2025geometry}, denoted RDO. Instead of taking a difference of means as in Basic RD, RDO parameterizes a single refusal direction per layer and optimizes it directly with gradient descent against a refusal-suppression objective on the harmful training split, exploiting differentiability through the model's hidden activations. We use the released implementation of \citet{wollschlager2025geometry} unchanged: the comparison against Basic RD isolates the contribution of learning the direction itself. As with Basic RD and Heretic, there is no per-prompt candidate bank and no learned aggregation $w$.

\paragraph{\new{RDO-CONE.}}
\new{\citet{wollschlager2025geometry} also argue that refusal is mediated by a
cone of directions rather than by a single line, and release a concept-cone
construction alongside the single-direction RDO objective. This is the prior
construction closest to the motivation of WRDO, since it likewise rejects the
assumption that one difference-of-means axis captures refusal, so we include it
as a separate baseline denoted RDO-CONE. It differs from WRDO in where the
multiplicity is placed: RDO-CONE learns a set of mutually constrained
directions that span a cone and ablates the whole cone, whereas WRDO keeps a
fixed bank of per-prompt directions and learns only how to aggregate them into
the composite of Equation~\ref{eq:rw}. We again use the released implementation
of \citet{wollschlager2025geometry} unchanged, with the authors' default cone
dimensionality.}

\subsection{WRD optimizers}
\label{sec:wrd_opt}

For optimizing the refusal-direction weights in the WRDO pipeline, we propose two distinct procedures. The first is a WRDO-tailored adaptation of separable CMA-ES, a well-known derivative-free optimizer suited to our low-dimensional setting. The second is a novel scheme based on group-relative policy optimization with importance sampling, designed specifically for the WRDO objective.

\paragraph{WRD-CMAES.}
CMA-ES \citep{hansen2001completely,hansen2016cmatutorial} is a standard derivative-free optimizer for continuous black-box objectives, well suited to the noisy and gradient-free nature of the WRDO loss. In our setting we apply it to $\mathcal{L}_J$ of Equation~\ref{eq:loss}: the search variables are the aggregation coefficients $w$ of the candidate bank, the search space is described in Appendix~\ref{app:constraints}, and each black-box query corresponds to one full WRDO evaluation, i.e.\ building the composite direction $r(w)$, intervening the model through the operator of Section~\ref{sec:operator}, and measuring the resulting ASR on the validation batch.
The concrete sampler choice, hyperparameter values and search-space bounds are listed in Appendix~\ref{sec:train_hyper}.

\new{\textbf{Relation to evolution strategies.} Both instantiations belong to the
evolution-strategy family
\citep{hansen2001completely,wierstra2014nes,salimans2017es}: a distribution over
$w$ is maintained, perturbed samples are scored by the black-box objective, and
the distribution is moved towards the better-scoring ones. WRD-CMAES is its
canonical member, and WRD-GRPO-IS reads as an ES with importance mixing, with
the Gaussian rollout of Equation~\ref{eq:rollout_w} as the search distribution
and the weights $\widetilde{v}_m$ as the mixing correction, so running both
compares ES against RL under one objective and one budget. The analogy stops at
the reward: NES-style updates see only the scalar fitness of each perturbation,
whereas WRD-GRPO-IS routes it through token-level policy ratios and can
therefore carry the KL term of Equation~\ref{eq:loss-grpo}.}

\paragraph{WRD-GRPO-IS.}
\label{sec:grpo}
The target loss function described in \ref{sec:objective} is a non-differentiable function, so it is natural to consider a reinforcement learning approach to its optimization. To keep the intervened model close to the clean base model on benign prompts, WRD-GRPO-IS augments the bare WRDO objective of Equation~\ref{eq:loss} with a KL-divergence penalty against the policy $\pi_{f_\theta}$ evaluated on a random batch from $\mathcal{D}_{\text{benign}}$, yielding the method-specific objective
\begin{equation}
\widetilde{\mathcal{L}}_J(w) = \mathcal{L}_J(w) + \beta\, \mathrm{KL}_w,
\label{eq:loss-grpo}
\end{equation}
with $\beta \geq 0$ and $\mathrm{KL}_w = \mathbb{E}_{x \sim \mathcal{D}_{\text{benign}}}\!\bigl[\mathrm{KL}\bigl(\pi_{T_{r(w)}(\theta)}(\cdot|x) \,\|\, \pi_{f_\theta}(\cdot|x)\bigr)\bigr]$ the KL divergence of the intervened policy from the clean policy on the benign batch. The KL term is specific to WRD-GRPO-IS: the WRD-CMAES sampler consumes the ASR-only signal of Equation~\ref{eq:loss} without any regularization.

\textbf{Group Relative Policy Optimization.}
Let us denote the language model parameterized by $\theta$ as a policy $\pi_\theta$.
Given a query $x$ drawn from a query set $\mathcal{D}$, it induces a distribution over responses $y$, whose likelihood factorizes as
\begin{align*}
\pi_\theta(y\mid x)=\prod_{t=1}^{|y|}\pi_\theta(y_t\mid x,y_{<t}),
\end{align*}
where $|y|$ denotes the number of tokens in $y$.
A query-response pair $(x,y)$ is scored by a reward model $\mathcal{S}$, yielding a per-prompt reward $\mathcal{S}(x,y) \in \{0,1\}$; in our setting $\mathcal{S}(x, y) = J(x, y)$ is the binary judge decision of Equation~\ref{eq:asr}.
The standard GRPO algorithm computes a relative advantage for each response within a group of $G$ responses generated for the same query from the old policy $\pi_{\theta_\text{old}}$.
Specifically, GRPO optimizes
\begin{equation}
\label{eq:grpo}
\mathcal{J}_\text{GRPO}(\theta) = \mathbb{E}_{x,\, \{y_i\} \sim \pi_{\theta_\text{old}}(\cdot \mid x)} \Bigg[
\frac{1}{G}\sum_{i=1}^{G}\frac{1}{|y_i|}\sum_{t=1}^{|y_i|}
\min\!\Big( \rho_{i,t}(\theta)\, \widehat{A}_{i}, \mathrm{clip}(\rho_{i,t}(\theta),\, 1-\varepsilon,\, 1+\varepsilon)\, \widehat{A}_{i} \Big) \Bigg],
\end{equation}
where the token-level importance ratio is
\begin{align*}
\rho_{i,t}(\theta)=\frac{\pi_\theta(y_{i,t}\mid x,y_{i,<t})}{\pi_{\theta_\text{old}}(y_{i,t}\mid x,y_{i,<t})},
\end{align*}
\new{We write importance ratios as $\rho$ throughout this section to keep them
distinct from the aggregation coefficients $w \in \mathbb{R}^N$ of
Section~\ref{sec:wdir}, which remain the only trainable quantity of WRDO.}
and all tokens in a response share the same group-relative advantage
\begin{align*}
\widehat{A}_{i}=\frac{\mathcal{S}(x,y_i)-\mathrm{mean}(\{\mathcal{S}(x,y_j)\}_{j=1}^{G})}{\mathrm{std}(\{\mathcal{S}(x,y_j)\}_{j=1}^{G})}.
\end{align*}

\textbf{Group Relative Policy Optimization with Importance Sampling.}
In the standard GRPO formulation \eqref{eq:grpo}, the group of responses for each query is sampled from the old policy $\pi_{\theta_\text{old}}$, and the clipped objective uses the token-level importance ratio between the current and old policies.
Here, we consider a stratified data-collection scheme. For each query $x$, we sample one response from each of $M$ behavior policies $\{\mu_m\}_{m=1}^M$ that are obtained by shifting the original model weights with formula \eqref{eq:operator}, i.e., $y_m \sim \mu_m(\cdot \mid x)$.
As the resulting group is off-policy with respect to $\pi_{\theta_\text{old}}$, directly applying \eqref{eq:grpo} would generally yield a biased estimate of the GRPO objective.
We introduce sequence-level importance sampling weights to correct this mismatch while retaining the original ratio against $\pi_{\theta_\text{old}}$.

Specifically, for each response $y_m$ sampled from $\mu_m$, we define the sequence-level importance weight
\begin{align*}
v_m = \frac{\pi_{\theta_\text{old}}(y_m \mid x)}{\mu_m(y_m \mid x)}.
\end{align*}
In practice, we stabilize the weights $\{v_m\}_{m=1}^M$ by applying clipping, denoted as $\widetilde{v}_m$, and treat them as constants with respect to $\theta$ during optimization.
This reweighting ensures that the overall objective remains consistent with the GRPO surrogate.

Analogous to GRPO, we assign a shared advantage to all tokens in a response.
However, to preserve the control-variate property under off-policy sampling, we compute the group baseline using the same sequence-level weights.
Given the group $\{y_m\}_{m=1}^M$ for query $x$, we first form the importance-weighted baseline
\begin{align*}
b(x)=\frac{\sum_{m=1}^{M} \widetilde{v}_m \, \mathcal{S}(x,y_m)}{\sum_{m=1}^{M} \widetilde{v}_m},
\end{align*}
and define the normalized group-relative advantage as
\begin{align*}
\widehat{A}_m
=
\frac{\mathcal{S}(x,y_m)-b(x)}
{\mathrm{std}\!\left(\{\mathcal{S}(x,y_j)\}_{j=1}^{M}\right)},
\end{align*}
where all tokens in $y_m$ share the same advantage, i.e., $\widehat{A}_{m,t}=\widehat{A}_m$ for all $t$.

Finally, our GRPO objective with importance sampling is
\begin{equation}
\label{eq:grpo_is_corrected}
\mathcal{J}_\text{GRPO-IS}(\theta) = \mathbb{E}_{x,\, y_m \sim \mu_m(\cdot \mid x)} \Bigg[ \frac{1}{M}\sum_{m=1}^{M} \widetilde{v}_m\, \frac{1}{|y_m|}\sum_{t=1}^{|y_m|}
\min\!\Big( \rho_{m,t}(\theta)\, \widehat{A}_{m}, \mathrm{clip}(\rho_{m,t}(\theta),\, 1-\varepsilon,\, 1+\varepsilon)\, \widehat{A}_{m} \Big) \Bigg],
\end{equation}
where the token-level importance ratio is defined as
\begin{align*}
\rho_{m,t}(\theta)=\frac{ \pi_{\theta} (y_{m,t} \mid x, y_{m,<t}) }{ \pi_{\theta_\text{old}} (y_{m,t} \mid x, y_{m,<t}) }.
\end{align*}
Compared with standard GRPO, our objective \eqref{eq:grpo_is_corrected} differs in that the group is collected by stratified sampling from multiple behavior policies $\{\mu_m\}$, and a sequence-level importance weight $\widetilde{v}_m$ is used to correct the off-policy sampling back to $\pi_{\theta_\text{old}}$.

In our setting, the trainable parameter is the coefficient vector $w \in \mathbb{R}^N$ of Section~\ref{sec:wdir} and the base policy $\pi_{\theta_\text{old}}$ is the clean model $f_\theta$. At each training step, the $M$ behavior policies $\{\mu_m\}_{m=1}^M$ are constructed by perturbing the current $w$ with Gaussian noise,
\begin{align}
\label{eq:rollout_w}
w_m = w + \varepsilon_m, \qquad \varepsilon_m \sim \mathcal{N}(0, \sigma^2 I_N),
\end{align}
with $\varepsilon_0 = 0$. Each $w_m$ defines a layerwise composite direction $r^{(l)}(w_m)$ via Equation~\ref{eq:rw} and an intervened model $\mu_m = T_{r(w_m)}(\theta_\text{old})$ via the operator of Section~\ref{sec:operator}. Only $w$ is updated by the surrogate of Equation~\ref{eq:grpo_is_corrected}, while the candidate bank $\mathcal{R}$, the base parameters $\theta_\text{old}$ and the schedule $\alpha^{(l)}$ are held fixed throughout training.

\section{Experimental Setup}
\label{sec:design}

\subsection{Data}
\label{sec:data}

The harmful pool is the union of six public benchmarks, namely AdvBench \citep{zou2023gcg}, MaliciousInstruct \citep{huang2023catastrophic}, the TDC2023 \citep{tdc2023}, HarmBench \citep{mazeika2024harmbench}, JailbreakBench \citep{chao2024jailbreakbench} and StrongREJECT \citep{souly2024strongreject}, deduplicated by exact prompt-string match and treated as a single distribution. The deduplicated pool is pre-partitioned into a training set, used only to compute the candidate bank, a validation set, used as the optimization objective in Equation~\ref{eq:asr}, and a test set, used only for the final evaluation. Concrete sampling seeds and the source of the prepared splits are documented in Appendix~\ref{app:splits}. The harmless baseline $\mathcal{D}_{\text{benign}}$ is a fixed-size sub-sample of Alpaca \citep{taori2023alpaca} used at bank-construction time.

\subsection{Models}
\label{sec:models}

We use three open-weight models: \texttt{Qwen/\allowbreak{}Qwen3-8B-Base} \citep{qwen3technicalreport}, \texttt{allenai/\allowbreak{}OLMo-3-1025-7B} \citep{olmo2025olmo3} and \texttt{Olmo-3-\allowbreak{}7B-Instruct} \citep{olmo2025olmo3}.
All evaluations use greedy decoding (\texttt{do\_sample}=\texttt{False}) with a maximum of $100$ new tokens per completion, one completion per prompt, the default system prompt of each model and no few-shot context.

\subsection{Judges}
\label{sec:judges}

We use three guard models, namely \texttt{allenai/wildguard} \citep{han2024wildguard}, \texttt{Qwen/\allowbreak{}Qwen3Guard-Gen-8B} \citep{qwen3guard} and \texttt{meta-llama/\allowbreak{}Llama-Guard-3-8B} \citep{inan2023llamaguard}. Each guard exposes its own decoding pipeline that maps a (prompt, response) pair to a binary safe/unsafe label, which we use directly as $J(x, y) \in \{0, 1\}$ in Equation~\ref{eq:asr}. The per-judge $\mathrm{ASR}_J \in [0, 1]$ is then obtained by averaging these per-prompt binary decisions over the test set and is reported separately for every judge. HarmBench \citep{mazeika2024harmbench} with the official classifier \texttt{cais/HarmBench-Llama-2-13b-cls} provides an additional external evaluator that is never used during optimization.

\subsection{Metrics}
\label{sec:metrics}

The primary metric is the $\mathrm{ASR}_J$ of Equation~\ref{eq:asr}, reported separately for each of the three judges $J$ on the held-out test set. The secondary metrics are the utility benchmarks of Section~\ref{sec:utility}, the HarmBench ASR of Section~\ref{sec:exp4}.

\subsection{Utility benchmarks}
\label{sec:utility}

\begin{table*}[!t]
\centering
\scriptsize
\setlength{\tabcolsep}{7pt}
\begin{tabular}{l l ccc ccc}
\toprule
& & \multicolumn{3}{c}{$\alpha=1.0$} & \multicolumn{3}{c}{$\alpha=3.0$} \\
\cmidrule(lr){3-5} \cmidrule(lr){6-8}
Model & Method & WG & Q3G & LG3 & WG & Q3G & LG3 \\
\midrule
\multirow{7}{*}{Qwen3-8B-Base}
 & No intervention  & 0.02 & 0.02 & 0.02 & 0.02 & 0.02 & 0.02 \\
 & Basic RD         & \new{0.22 {\tiny$\pm$0.02}} & \new{0.11 {\tiny$\pm$0.03}} & \new{0.44 {\tiny$\pm$0.02}} & \new{0.81 {\tiny$\pm$0.03}} & \new{0.80 {\tiny$\pm$0.03}} & \new{0.91 {\tiny$\pm$0.04}} \\
 & Heretic          & \new{0.77 {\tiny$\pm$0.04}} & \new{0.83 {\tiny$\pm$0.05}} & \new{0.81 {\tiny$\pm$0.05}} & \new{0.77 {\tiny$\pm$0.04}} & \new{0.83 {\tiny$\pm$0.05}} & \new{0.81 {\tiny$\pm$0.05}} \\
 & RDO              & \new{0.34 {\tiny$\pm$0.05}} & \new{0.41 {\tiny$\pm$0.04}} & \new{0.48 {\tiny$\pm$0.05}} & \new{0.34 {\tiny$\pm$0.05}} & \new{0.41 {\tiny$\pm$0.04}} & \new{0.48 {\tiny$\pm$0.05}} \\
 & Uniform-avg WRD  & \new{0.08 {\tiny$\pm$0.001}} & \new{0.25 {\tiny$\pm$0.01}} & \new{0.32 {\tiny$\pm$0.02}} & \new{0.002 {\tiny$\pm$0.0003}} & \new{0.89 {\tiny$\pm$0.03}} & \new{0.92 {\tiny$\pm$0.02}} \\
 & WRD-CMAES        & \new{0.88 {\tiny$\pm$0.05}} & \new{0.83 {\tiny$\pm$0.02}} & \new{\textbf{0.89} {\tiny$\pm$0.02}} & \new{0.91 {\tiny$\pm$0.05}} & \new{\textbf{0.99} {\tiny$\pm$0.002}} & \new{\textbf{0.98} {\tiny$\pm$0.002}} \\
 & WRD-GRPO-IS      & \new{\textbf{0.91} {\tiny$\pm$0.07}} & \new{\textbf{0.94} {\tiny$\pm$0.02}} & \new{0.87 {\tiny$\pm$0.02}} & \new{\textbf{0.93} {\tiny$\pm$0.09}} & \new{0.95 {\tiny$\pm$0.02}} & \new{0.96 {\tiny$\pm$0.01}} \\
\midrule
\multirow{7}{*}{OLMo-3-1025-7B}
 & No intervention  & 0.08 & 0.09 & 0.07 & 0.08 & 0.09 & 0.07 \\
 & Basic RD         & \new{0.41 {\tiny$\pm$0.02}} & \new{0.51 {\tiny$\pm$0.03}} & \new{0.43 {\tiny$\pm$0.03}} & \new{0.76 {\tiny$\pm$0.02}} & \new{0.72 {\tiny$\pm$0.03}} & \new{0.73 {\tiny$\pm$0.02}} \\
 & Heretic          & \new{0.76 {\tiny$\pm$0.03}} & \new{0.73 {\tiny$\pm$0.04}} & \new{0.70 {\tiny$\pm$0.03}} & \new{0.76 {\tiny$\pm$0.03}} & \new{0.73 {\tiny$\pm$0.04}} & \new{0.70 {\tiny$\pm$0.03}} \\
 & RDO              & \new{0.61 {\tiny$\pm$0.04}} & \new{0.67 {\tiny$\pm$0.03}} & \new{0.73 {\tiny$\pm$0.03}} & \new{0.61 {\tiny$\pm$0.04}} & \new{0.67 {\tiny$\pm$0.03}} & \new{0.73 {\tiny$\pm$0.03}} \\
 & Uniform-avg WRD  & \new{0.64 {\tiny$\pm$0.02}} & \new{0.66 {\tiny$\pm$0.01}} & \new{0.68 {\tiny$\pm$0.02}} & \new{0.77 {\tiny$\pm$0.03}} & \new{0.76 {\tiny$\pm$0.02}} & \new{0.79 {\tiny$\pm$0.01}} \\
 & WRD-CMAES        & \new{\textbf{0.87} {\tiny$\pm$0.02}} & \new{\textbf{0.86} {\tiny$\pm$0.03}} & \new{\textbf{0.82} {\tiny$\pm$0.02}} & \new{0.88 {\tiny$\pm$0.01}} & \new{0.88 {\tiny$\pm$0.02}} & \new{0.85 {\tiny$\pm$0.03}} \\
 & WRD-GRPO-IS      & \new{0.82 {\tiny$\pm$0.02}} & \new{\textbf{0.86} {\tiny$\pm$0.03}} & \new{0.79 {\tiny$\pm$0.02}} & \new{\textbf{0.92} {\tiny$\pm$0.02}} & \new{\textbf{0.97} {\tiny$\pm$0.03}} & \new{\textbf{0.97} {\tiny$\pm$0.02}} \\
\midrule
\multirow{7}{*}{Olmo-3-7B-Instruct}
 & No intervention  & \new{0.003} & \new{0.003} & 0.004 & \new{0.003} & \new{0.003} & 0.004 \\
 & Basic RD         & \new{0.40 {\tiny$\pm$0.02}} & \new{0.42 {\tiny$\pm$0.01}} & \new{0.43 {\tiny$\pm$0.03}} & \new{0.53 {\tiny$\pm$0.02}} & \new{0.61 {\tiny$\pm$0.03}} & \new{0.59 {\tiny$\pm$0.01}} \\
 & Heretic          & \new{0.62 {\tiny$\pm$0.03}} & \new{0.58 {\tiny$\pm$0.04}} & \new{0.69 {\tiny$\pm$0.04}} & \new{0.62 {\tiny$\pm$0.03}} & \new{0.58 {\tiny$\pm$0.04}} & \new{0.69 {\tiny$\pm$0.04}} \\
 & RDO              & \new{0.40 {\tiny$\pm$0.05}} & \new{0.52 {\tiny$\pm$0.04}} & \new{0.53 {\tiny$\pm$0.04}} & \new{0.40 {\tiny$\pm$0.05}} & \new{0.52 {\tiny$\pm$0.04}} & \new{0.53 {\tiny$\pm$0.04}} \\
 & Uniform-avg WRD  & \new{0.49 {\tiny$\pm$0.02}} & \new{0.42 {\tiny$\pm$0.01}} & \new{0.45 {\tiny$\pm$0.03}} & \new{0.62 {\tiny$\pm$0.02}} & \new{0.67 {\tiny$\pm$0.01}} & \new{0.67 {\tiny$\pm$0.01}} \\
 & WRD-CMAES        & \new{\textbf{0.92} {\tiny$\pm$0.02}} & \new{\textbf{0.89} {\tiny$\pm$0.01}} & \new{\textbf{0.93} {\tiny$\pm$0.01}} & \new{0.89 {\tiny$\pm$0.02}} & \new{\textbf{0.93} {\tiny$\pm$0.02}} & \new{0.86 {\tiny$\pm$0.01}} \\
 & WRD-GRPO-IS      & \new{\textbf{0.92} {\tiny$\pm$0.01}} & \new{0.88 {\tiny$\pm$0.02}} & \new{\textbf{0.93} {\tiny$\pm$0.02}} & \new{\textbf{0.93} {\tiny$\pm$0.01}} & \new{0.90 {\tiny$\pm$0.02}} & \new{\textbf{0.88} {\tiny$\pm$0.01}} \\
\bottomrule
\end{tabular}
\caption{Attack setup. Test ASR (in $[0, 1]$) on held-out harmful prompts, reported at the two extremes of the intensity grid, $\alpha \in \{1.0, 3.0\}$. Judges: WG = WildGuard, Q3G = Qwen3Guard-Gen-8B, LG3 = Llama-Guard-3-8B. Rows for $\alpha$-independent methods are repeated across columns for layout consistency. \textbf{Bold} marks the best result in each column within each model block. Higher is a stronger attack. \new{The intermediate intensities $\alpha \in \{1.5, 2.0\}$ are reported in Table~\ref{tab:exp1-main-app} of Appendix~\ref{app:attack-intermediate}. Every intervened cell is the mean over three independent repetitions of the full pipeline, with the standard deviation over those repetitions. Cells that would round to $0.00$ are reported at three decimals; none is an exact zero.}}
\label{tab:exp1-main}
\end{table*}

Utility preservation is evaluated through the Language Model Evaluation Harness \citep{eval-harness} on \texttt{arc\_easy} and \texttt{arc\_challenge} \citep{clark2018arc}, \texttt{mmlu} \citep{hendrycks2020mmlu}, \texttt{gsm8k} and word-level perplexity on \texttt{wikitext}. Accuracy and perplexity are reported separately, with the direction of improvement made explicit in the tables.
\section{Results}
\label{sec:results}

\subsection{Attack setup}
\label{sec:exp1}

Table~\ref{tab:exp1-main} reports the test-set ASR of every method on the three base models, evaluated separately against the three guard judges, at the two extremes of the peak intensity grid $\alpha \in \{1.0, 3.0\}$ of the layerwise schedule introduced in Section~\ref{sec:operator} (Equation~\ref{eq:operator}; full form in Appendix~\ref{app:schedule}). \new{The intermediate intensities $\alpha \in \{1.5, 2.0\}$ are deferred to Table~\ref{tab:exp1-main-app} in Appendix~\ref{app:attack-intermediate}, where the same conclusions hold.} The no-intervention rows provide the absence-of-attack baselines on the same test set. The Heretic entries are duplicated across the $\alpha$ columns because Heretic does not consume this hyperparameter from the outside: it selects its own shift coefficient internally through the toolkit's Bayesian search, so a single value is reported per model and judge and repeated across the $\alpha$ grid for layout consistency. The RDO rows are duplicated for the same reason: RDO does not expose $\alpha$ as a hyperparameter, the intensity is baked into the gradient-learned direction itself, and the resulting per-model, per-judge value is repeated across the $\alpha$ grid.

To check whether the apparent benefit of WRDO is driven by the per-prompt candidate bank alone or by the learned aggregation on top of it, we introduce the Uniform-avg WRD sanity-check baseline in this experiment. Uniform-avg WRD uses the same candidate bank of Section~\ref{sec:bank} as the learned WRDO instantiations, but fixes the coefficients to the uniform vector $w_i = 1/N$ without any optimization.

\textbf{Discussion.} First, in nearly every configuration the two learned WRDO instantiations (WRD-CMAES and WRD-GRPO-IS) outperform every prior single-direction baseline by a clear margin, establishing the superiority of the weighted-aggregation approach over the difference-of-means recipe, the Heretic toolkit and the gradient-learned RDO baseline. Second, the Uniform-avg WRD baseline is consistently dominated by the learned WRDO instantiations across the $\alpha$ grid, confirming that learning the aggregation coefficients $w$ is what drives the gain, not the per-prompt bank alone.

\subsection{\new{Concept cones and scale}}
\label{sec:cone}

\new{Two comparisons fall outside the grid of Table~\ref{tab:exp1-main} and are
reported in Appendix~\ref{app:extra}. Against the concept cone of
\citet{wollschlager2025geometry}, which treats refusal as a subspace rather than
a line, the two are close: RDO-CONE leads by $3.3$ ASR points on
Qwen3-8B-Base, WRD-GRPO-IS by $5.5$ and $8.5$ on the two OLMo checkpoints. Repeating
the attack setup on the $21$B \texttt{openai/gpt-oss-20b} \citep{gptoss2025}
reproduces the ranking with wider margins.}

\begin{table*}[t]
\centering
\scriptsize
\setlength{\tabcolsep}{2pt}
\resizebox{\linewidth}{!}{%
\begin{tabular}{l l ccccc ccccc}
\toprule
& & \multicolumn{5}{c}{$\alpha=1.0$} & \multicolumn{5}{c}{$\alpha=3.0$} \\
\cmidrule(lr){3-7} \cmidrule(lr){8-12}
Model & Method & ARC-e & ARC-c & MMLU & GSM8K & PPL & ARC-e & ARC-c & MMLU & GSM8K & PPL \\
\midrule
\multirow{6}{*}{Qwen3-8B-Base}
 & No intervention & 0.63 $\pm$ 0.01 & 0.38 $\pm$ 0.01 & 0.67 $\pm$ 0.004 & 0.84 $\pm$ 0.01 & 11.11 & 0.63 $\pm$ 0.01 & 0.38 $\pm$ 0.01 & 0.67 $\pm$ 0.004 & 0.84 $\pm$ 0.01 & 11.11 \\
 & Basic RD        & 0.59 $\pm$ 0.01 & 0.36 $\pm$ 0.01 & 0.64 $\pm$ 0.004 & 0.67 $\pm$ 0.01 & 13.81 & 0.26 $\pm$ 0.01 & 0.23 $\pm$ 0.01 & 0.23 $\pm$ 0.004 & 0.00 $\pm$ 0.00 & $>10^{12}$ \\
 & Heretic         & 0.51 $\pm$ 0.01 & 0.30 $\pm$ 0.01 & 0.27 $\pm$ 0.004 & 0.19 $\pm$ 0.01 & 13.44 & 0.51 $\pm$ 0.01 & 0.30 $\pm$ 0.01 & 0.27 $\pm$ 0.004 & 0.19 $\pm$ 0.01 & 13.44 \\
 & RDO             & 0.63 $\pm$ 0.01 & 0.37 $\pm$ 0.01 & 0.67 $\pm$ 0.004 & 0.82 $\pm$ 0.01 & 11.16 & 0.63 $\pm$ 0.01 & 0.37 $\pm$ 0.01 & 0.67 $\pm$ 0.004 & 0.82 $\pm$ 0.01 & 11.16 \\
 & WRD-CMAES       & 0.53 $\pm$ 0.01 & 0.34 $\pm$ 0.01 & 0.67 $\pm$ 0.003 & 0.01 $\pm$ 0.003 & 161.18 & 0.37 $\pm$ 0.01 & 0.29 $\pm$ 0.01 & 0.51 $\pm$ 0.004 & 0.004 $\pm$ 0.002 & $>10^{4}$ \\
 & WRD-GRPO-IS     & 0.63 $\pm$ 0.01 & 0.38 $\pm$ 0.01 & 0.65 $\pm$ 0.004 & 0.83 $\pm$ 0.01 & 11.15 & 0.64 $\pm$ 0.01 & 0.38 $\pm$ 0.01 & 0.62 $\pm$ 0.004 & 0.83 $\pm$ 0.01 & 11.35 \\
\midrule
\multirow{6}{*}{OLMo-3-1025-7B}
 & No intervention & 0.79 $\pm$ 0.01 & 0.48 $\pm$ 0.015 & 0.62 $\pm$ 0.004 & 0.53 $\pm$ 0.014 & 62.5 & 0.79 $\pm$ 0.01 & 0.48 $\pm$ 0.015 & 0.62 $\pm$ 0.004 & 0.53 $\pm$ 0.014 & 62.5 \\
 & Basic RD        & 0.80 $\pm$ 0.01 & 0.48 $\pm$ 0.01 & 0.61 $\pm$ 0.004 & 0.53 $\pm$ 0.01 & 62.71 & 0.42 $\pm$ 0.01 & 0.26 $\pm$ 0.01 & 0.32 $\pm$ 0.004 & 0.0008 $\pm$ 0.0008 & 432.02 \\
 & Heretic         & 0.79 $\pm$ 0.01 & 0.48 $\pm$ 0.01 & 0.62 $\pm$ 0.004 & 0.52 $\pm$ 0.01 & 62.61 & 0.79 $\pm$ 0.01 & 0.48 $\pm$ 0.01 & 0.62 $\pm$ 0.004 & 0.52 $\pm$ 0.01 & 62.61 \\
 & RDO             & 0.80 $\pm$ 0.01 & 0.48 $\pm$ 0.01 & 0.62 $\pm$ 0.004 & 0.56 $\pm$ 0.01 & 63.32 & 0.80 $\pm$ 0.01 & 0.48 $\pm$ 0.01 & 0.62 $\pm$ 0.004 & 0.56 $\pm$ 0.01 & 63.32 \\
 & WRD-CMAES       & 0.79 $\pm$ 0.01 & 0.48 $\pm$ 0.01 & 0.62 $\pm$ 0.004 & 0.52 $\pm$ 0.01 & 62.40 & 0.79 $\pm$ 0.01 & 0.48 $\pm$ 0.01 & 0.617 $\pm$ 0.004 & 0.51 $\pm$ 0.01 & 61.70 \\
 & WRD-GRPO-IS     & 0.79 $\pm$ 0.01 & 0.48 $\pm$ 0.015 & 0.62 $\pm$ 0.004 & 0.53 $\pm$ 0.014 & 62.5 & 0.78 $\pm$ 0.01 & 0.49 $\pm$ 0.015 & 0.61 $\pm$ 0.004 & 0.55 $\pm$ 0.014 & 62.21 \\
\midrule
\multirow{6}{*}{Olmo-3-7B-Instruct}
 & No intervention & 0.76 $\pm$ 0.01 & 0.48 $\pm$ 0.01 & 0.53 $\pm$ 0.004 & 0.67 $\pm$ 0.01 & 100.83 & 0.76 $\pm$ 0.01 & 0.48 $\pm$ 0.01 & 0.53 $\pm$ 0.004 & 0.67 $\pm$ 0.01 & 100.83 \\
 & Basic RD        & 0.77 $\pm$ 0.01 & 0.48 $\pm$ 0.01 & 0.54 $\pm$ 0.004 & 0.67 $\pm$ 0.01 & 100.81 & 0.42 $\pm$ 0.01 & 0.31 $\pm$ 0.01 & 0.17 $\pm$ 0.004 & 0.01 $\pm$ 0.01 & $>10^5$ \\
 & Heretic         & 0.76 $\pm$ 0.01 & 0.48 $\pm$ 0.01 & 0.54 $\pm$ 0.004 & 0.67 $\pm$ 0.01 & 101.27 & 0.76 $\pm$ 0.01 & 0.48 $\pm$ 0.01 & 0.54 $\pm$ 0.004 & 0.67 $\pm$ 0.01 & 101.27 \\
 & RDO             & 0.76 $\pm$ 0.01 & 0.48 $\pm$ 0.01 & 0.53 $\pm$ 0.004 & 0.65 $\pm$ 0.01 & 101.34 & 0.76 $\pm$ 0.01 & 0.48 $\pm$ 0.01 & 0.53 $\pm$ 0.004 & 0.65 $\pm$ 0.01 & 101.34 \\
 & WRD-CMAES       & 0.76 $\pm$ 0.01 & 0.48 $\pm$ 0.01 & 0.53 $\pm$ 0.004 & 0.68 $\pm$ 0.01 & 101.19 & 0.74 $\pm$ 0.01 & 0.46 $\pm$ 0.01 & 0.50 $\pm$ 0.004 & 0.64 $\pm$ 0.01 & 118.19 \\
 & WRD-GRPO-IS     & 0.76 $\pm$ 0.01 & 0.48 $\pm$ 0.01 & 0.53 $\pm$ 0.004 & 0.67 $\pm$ 0.01 & 100.81 & 0.76 $\pm$ 0.01 & 0.47 $\pm$ 0.01 & 0.53 $\pm$ 0.004 & 0.67 $\pm$ 0.01 & 100.49 \\
\bottomrule
\end{tabular}%
}
\caption{Utility preservation per base model, reported at $\alpha \in \{1.0, 3.0\}$. Higher is better for ARC-e, ARC-c, MMLU and GSM8K (accuracy as a fraction in $[0, 1]$); lower is better for PPL (WikiText word-level perplexity). Rows for $\alpha$-independent methods are repeated across columns for layout consistency. Intermediate values $\alpha \in \{1.5, 2.0\}$ are reported in Table~\ref{tab:exp2-utility-app} of Appendix~\ref{app:utility-intermediate}.}
\label{tab:exp2-utility}
\end{table*}

\subsection{Utility preservation}
\label{sec:exp2}

Table~\ref{tab:exp2-utility} reports utility benchmarks for the same $w^\star$ that produced Table~\ref{tab:exp1-main}, per base model, at the two extremes of the intensity grid $\alpha \in \{1.0, 3.0\}$. Intermediate values $\alpha \in \{1.5, 2.0\}$ are deferred to Table~\ref{tab:exp2-utility-app} in Appendix~\ref{app:utility-intermediate}. In this experiment we restrict attention to the WildGuard-optimized runs only: $w^\star$ in every cell below is the coefficient vector obtained by optimizing against WildGuard. Accuracy is reported for ARC-easy (ARC-e), ARC-challenge (ARC-c), MMLU and GSM8K; word-level perplexity is reported for WikiText (PPL). As in Section~\ref{sec:exp1}, the Heretic entries are duplicated across the $\alpha$ columns because Heretic selects its own shift coefficient internally rather than consuming $\alpha$ from the outside, so a single value per model and metric is repeated across the $\alpha$ grid for layout consistency.

\textbf{Discussion.} First, the prior single-direction baselines degrade general capabilities substantially at the larger values of $\alpha$: accuracy on ARC and GSM8K collapses and WikiText perplexity grows by orders of magnitude. \new{Heretic is partially protected, since its Bayesian search minimizes a KL
divergence to the unintervened model, and it does hold perplexity near the clean
value throughout. It is not uniformly utility-preserving, however: on
Qwen3-8B-Base it costs $40$ MMLU and $65$ GSM8K points, so its KL target
constrains the benign output distribution without protecting downstream
benchmarks.} Second, among the learned WRDO instantiations WRD-CMAES exhibits the same degradation pattern as the prior baselines: the Optuna target consumed by the sampler is the bare ASR-only signal without a KL term, so the optimizer is free to find directions that succeed at suppressing refusal but disrupt unrelated capabilities. Third, WRD-GRPO-IS preserves utility best because its surrogate of Equation~\ref{eq:grpo_is_corrected} optimizes the KL-regularized objective of Equation~\ref{eq:loss-grpo}, keeping the edit close to the clean policy even at high $\alpha$.

\subsection{HarmBench}
\label{sec:exp4}

We additionally re-evaluate the considered methods through an external benchmark that was not used during optimization, in order to corroborate the significance of the attack-setup results with a training-independent evaluator and to control for potential overfitting to the guard judges of Section~\ref{sec:judges}. Because HarmBench is designed primarily as an evaluator of instruction-tuned models, in this setup we restrict attention to the instruct variants of our model pool, additionally including \texttt{Qwen/Qwen3-8B} \citep{qwen3technicalreport}.

\begin{table}[t]
\centering
\small
\setlength{\tabcolsep}{6pt}
\begin{tabular}{l cccc cccc}
\toprule
& \multicolumn{4}{c}{Qwen/Qwen3-8B} & \multicolumn{4}{c}{Olmo-3-7B-Instruct} \\
\cmidrule(lr){2-5} \cmidrule(lr){6-9}
Method & 1.0 & 1.5 & 2.0 & 3.0 & 1.0 & 1.5 & 2.0 & 3.0 \\
\midrule
No intervention & 0.07 & 0.07 & 0.07 & 0.07 & 0.00 & 0.00 & 0.00 & 0.00 \\
\midrule
Basic RD    & 0.22 & 0.31 & 0.46 & 0.48 & 0.51 & 0.77 & 0.68 & 0.66 \\
Heretic         & 0.87 & 0.87 & 0.87 & 0.87 & 0.77 & 0.77 & 0.77 & 0.77 \\
RDO             & 0.39 & 0.39 & 0.39 & 0.39 & 0.20 & 0.20 & 0.20 & 0.20 \\
WRD-CMAES       & 0.87 & 0.92 & \textbf{0.97} & 0.82 & 0.81 & 0.85 & \textbf{0.88} & 0.89 \\
WRD-GRPO-IS     & \textbf{0.95} & \textbf{0.94} & 0.91 & \textbf{0.97} & \textbf{0.83} & \textbf{0.93} & 0.86 & \textbf{0.92} \\
\bottomrule
\end{tabular}
\caption{HarmBench Standard ASR (in $[0, 1]$) computed with \texttt{cais/HarmBench-Llama-2-13b-cls}, varying $\alpha$ and model. \textbf{Bold} marks the best result in each column within each model block. Higher is a stronger attack.}
\label{tab:exp4-harmbench}
\end{table}

\begin{table*}[!t]
\centering
\small
\setlength{\tabcolsep}{3pt}
\resizebox{\linewidth}{!}{%
\begin{tabular}{l ccc ccc ccc ccc}
\toprule
& \multicolumn{3}{c}{$\alpha = -1.0$} & \multicolumn{3}{c}{$\alpha = -1.5$} & \multicolumn{3}{c}{$\alpha = -2.0$} & \multicolumn{3}{c}{$\alpha = -3.0$} \\
\cmidrule(lr){2-4} \cmidrule(lr){5-7} \cmidrule(lr){8-10} \cmidrule(lr){11-13}
Method & WG & Q3G & LG3 & WG & Q3G & LG3 & WG & Q3G & LG3 & WG & Q3G & LG3 \\
\midrule
Abliterated (no defense) & 0.85 & 0.83 & 0.84 & 0.85 & 0.83 & 0.83 & 0.85 & 0.83 & 0.84 & 0.85 & 0.83 & 0.84 \\
\midrule
Basic RD & \new{0.84 {\tiny$\pm$0.02}} & \new{0.85 {\tiny$\pm$0.03}} & \new{0.82 {\tiny$\pm$0.02}} & \new{0.79 {\tiny$\pm$0.03}} & \new{0.84 {\tiny$\pm$0.02}} & \new{0.79 {\tiny$\pm$0.03}} & \new{0.80 {\tiny$\pm$0.03}} & \new{0.87 {\tiny$\pm$0.02}} & \new{0.87 {\tiny$\pm$0.03}} & \new{0.63 {\tiny$\pm$0.02}} & \new{0.68 {\tiny$\pm$0.03}} & \new{0.74 {\tiny$\pm$0.02}} \\
WRD-CMAES & \new{\textbf{0.47} {\tiny$\pm$0.05}} & \new{0.56 {\tiny$\pm$0.04}} & \new{0.53 {\tiny$\pm$0.06}} & \new{\textbf{0.011} {\tiny$\pm$0.004}} & \new{\textbf{0.023} {\tiny$\pm$0.005}} & \new{0.073 {\tiny$\pm$0.004}} & \new{0.021 {\tiny$\pm$0.005}} & \new{0.053 {\tiny$\pm$0.006}} & \new{0.042 {\tiny$\pm$0.004}} & \new{\textbf{0.014} {\tiny$\pm$0.005}} & \new{0.086 {\tiny$\pm$0.006}} & \new{0.079 {\tiny$\pm$0.008}} \\
WRD-GRPO-IS & \new{0.50 {\tiny$\pm$0.08}} & \new{\textbf{0.52} {\tiny$\pm$0.07}} & \new{\textbf{0.47} {\tiny$\pm$0.09}} & \new{0.014 {\tiny$\pm$0.007}} & \new{0.034 {\tiny$\pm$0.008}} & \new{\textbf{0.047} {\tiny$\pm$0.006}} & \new{\textbf{0.015} {\tiny$\pm$0.007}} & \new{\textbf{0.014} {\tiny$\pm$0.003}} & \new{\textbf{0.020} {\tiny$\pm$0.005}} & \new{0.019 {\tiny$\pm$0.009}} & \new{\textbf{0.015} {\tiny$\pm$0.003}} & \new{\textbf{0.022} {\tiny$\pm$0.008}} \\
\bottomrule
\end{tabular}%
}
\caption{Defense setup. Test ASR (in $[0, 1]$) on the abliterated checkpoint, across the sign-reversed intensity $\alpha$ and the three judges. Lower is a stronger defense. \new{The three intervened rows are means over three independently optimized $w^\star$, with the standard deviation over those runs; the Abliterated row receives no intervention and is reported from a single evaluation. \textbf{Bold} marks the strongest defense per column.}}
\label{tab:exp3-defense}
\end{table*}

Table~\ref{tab:exp4-harmbench} reports HarmBench ASR. We adopt the same HarmBench evaluation setup as \citep{arditi2024refusal}: we take the prompt set from \citep{mazeika2024harmbench} and score the responses generated by the intervened model with the official HarmBench classifier \texttt{cais/HarmBench-Llama-2-13b-cls}. As in the utility experiment of Section~\ref{sec:exp2}, we restrict this evaluation to the WildGuard-optimized runs: for every method the intervened model evaluated by HarmBench is the one whose coefficient vector was selected against WildGuard during training. HarmBench itself is never used during optimization and serves as an external evaluator that controls for judge-specific overfitting.

\textbf{Discussion.} The HarmBench evaluation confirms the superiority of the WRDO approach observed in Section~\ref{sec:exp1}: the learned WRDO instantiations attain the highest HarmBench ASR across the $\alpha$ grid. The agreement between the guard-judge metric and a classifier that was never used during optimization confirms that the WRDO advantage over the prior single-direction recipes is a real effect rather than an artifact of the training signal.

\subsection{Defense setup}
\label{sec:exp3}

We additionally consider a defense setup that reverses the role of WRDO: instead of unlocking harmful behavior on an aligned model, we attempt to suppress it on a model whose refusal has already been broken. Notably, we operate on a third-party abliterated checkpoint whose editing procedure is unknown to us, so we work exclusively from its weights without any knowledge of how the refusal removal was performed.

Table~\ref{tab:exp3-defense} reports the defense setup. As the abliterated checkpoint we take the publicly released \texttt{Orenguteng/\allowbreak{}Llama-3.1-8B-\allowbreak{}Lexi-\allowbreak{}Uncensored-V2}, an uncensored derivative of Llama-3.1-8B-Instruct whose refusal behavior has already been removed by a third-party procedure. We apply the same WRDO pipeline to this checkpoint with the same $\alpha$ grid as in the attack setup, only with reversed sign. For Basic RD this coincides exactly with the defense recipe of Section~3.2 of \citep{arditi2024refusal}. The Heretic toolkit and RDO do not expose a defense mode and are therefore excluded from this experiment.

\textbf{Discussion.} The two learned WRDO instantiations clearly win the defense setup, driving the abliterated checkpoint's ASR far below the no-defense baseline, while Basic RD fails to restore refusal. This shows that the proposed methods adapt to the alternative setup through a simple sign flip, with no change to the optimizer or the candidate bank.

\subsection{Cross-judge transfer}
\label{sec:cross}

A cross-judge transfer analysis (Appendix~\ref{app:cross}) identifies WildGuard as the most reliable optimization signal among the considered guards: interventions optimized against WildGuard transfer well to the other two judges, while the reverse direction is less stable.

\section{Conclusion}
\label{sec:conclusion}

We introduced Weighted Refusal Direction Optimization (WRDO), a procedure that constructs a refusal direction by learning aggregation coefficients $w$ over a bank of per-prompt directions, and we presented two derivative-free instantiations of the resulting black-box problem: a WRDO-tailored adaptation of separable CMA-ES and a proposed GRPO variant with importance sampling.

Across three open-weight models, three guard judges and the HarmBench classifier, the two WRDO instantiations exceed all prior single-direction baselines and transfer cleanly to the defense setup. Among them, WRD-GRPO-IS is the most reliable: it preserves the general capabilities of the intervened model substantially better than every other method considered, and the cross-judge analysis singles out WildGuard as the most reliable optimization signal, so we recommend WRD-GRPO-IS optimized against WildGuard as the default configuration for both auditing and defense.

\section{Limitations}
\label{sec:limitations}

The full protocol of Sections~\ref{sec:exp1}--\ref{sec:exp3} covers three open-weight checkpoints of comparable size, in the $7$--$8$B parameter range.

A second limitation concerns the WRD-CMAES instantiation: its current implementation consumes the bare ASR-only objective of Equation~\ref{eq:loss} and does not yet support the KL regularizer of Equation~\ref{eq:loss-grpo}.
Extending the Optuna interface to expose a KL-aware sampling target is under active development, and we expect to incorporate this support in a future revision of the codebase.

\section{Acknowledgments}
The work of G. Evseev, D. Bylinkin, D. Medyakov, V. Aletov, A. Beznosikov was supported by the Ministry of Economic Development of the Russian Federation (agreement No. 139-15-2025-013, dated June 20, 2025, IGK 000000C313925P4B0002).

\section{Ethical Considerations}
\label{sec:ethics}

This work is inherently dual-use: methods for identifying and modifying refusal directions can help analyze and improve model safety, but they can also be used to weaken safeguards in aligned models. In particular, interventions that suppress refusal may increase the risk of harmful compliance in domains such as cyber abuse, privacy violations, disinformation, or other unsafe requests.

We therefore view WRDO primarily as a tool for controlled mechanistic analysis and safety evaluation. Our experiments are conducted offline on open-weight models and use automated guard judges, HarmBench-style evaluation, and utility benchmarks to study both safety and capability effects. The goal is to better understand refusal representations, not to encourage unsafe deployment or misuse.

The broader implication of our findings is that refusal behavior may be concentrated in low-dimensional internal features and can therefore be brittle under targeted interventions. This suggests that refusal alone should not be treated as a complete safety mechanism; robust deployment requires layered defenses, including safer training, adversarial evaluation, monitoring, external classifiers, and system-level policy enforcement.

At the same time, the defense experiment shows a constructive use of the same machinery: WRDO-like methods may help diagnose and partially restore refusal behavior in modified or abliterated checkpoints. Overall, we hope this work supports more robust alignment methods and encourages careful, transparent evaluation of both the risks and defensive uses of refusal-direction interventions.

\end{mainpart}

\begin{appendixpart}

\section{Coefficient-Vector Constraints}
\label{app:constraints}

The coefficients $w_i$ in Equation~\ref{eq:rw} are signed real numbers, with no further normalization across coordinates beyond the per-layer $L_2$ renormalization already built into Equation~\ref{eq:rw}. The feasible set $\mathcal{W}$ of Equation~\ref{eq:attack} is the per-coordinate box
\begin{equation}
\mathcal{W} = [w_{\min}, w_{\max}]^N,
\label{eq:feasible}
\end{equation}
with bounds $w_{\min}$ and $w_{\max}$ fixed as hyperparameters of the search and shared across all derivative-free methods of Section~\ref{sec:methods}. We do not impose a simplex constraint, a unit-norm constraint or a top-$k$ sparsity constraint: the box of Equation~\ref{eq:feasible} is the only restriction on $w$. All derivative-free methods are initialized at the uniform-average point $w_i = 1/N$, which coincides with the Uniform-avg WRD sanity-check baseline of Section~\ref{sec:exp1}.

\section{Layerwise Intensity Schedule}
\label{app:schedule}

The fixed layerwise schedule $\alpha^{(l)}$ of Equation~\ref{eq:operator} is adopted verbatim from the Heretic toolkit \citep{heretic}. Let $\ell^\star = \texttt{max\_weight\_position} \cdot (L-1)$ denote the absolute peak position in the stack, $\Delta = \texttt{min\_weight\_distance} \cdot (L-1)$ the half-width of the active window, $\alpha_{\min} = \texttt{min\_weight}$ the boundary value of the schedule, and $d_l = |l - \ell^\star| / \Delta$ the normalized distance of layer $l$ from the peak. The schedule is
\begin{equation}
\alpha^{(l)} = \begin{cases}
\alpha + d_l\,(\alpha_{\min} - \alpha), & d_l \leq 1, \\
0, & \text{otherwise},
\end{cases}
\label{eq:schedule}
\end{equation}
so that the intensity decays linearly from the peak value $\alpha = \texttt{max\_weight}$ at $\ell^\star$ to $\alpha_{\min}$ at the window boundary, and no intervention is applied to layers further than $\Delta$ from the peak. We follow the parameter naming of \citep{heretic} verbatim. Throughout the paper we use $\texttt{max\_weight\_position} = 0.7$, $\texttt{min\_weight\_distance} = 0.3$ and $\alpha_{\min} = 0$, leaving $\alpha = \texttt{max\_weight} \in \{1.0, 1.5, 2.0, 3.0\}$ as the only varied parameter.

\section{Data Splits and Sampling Seeds}
\label{app:splits}

\paragraph{Source of the splits.}
All harmful-prompt splits are loaded from the prepared dataset \texttt{erinya/dataset\_splits} on the Hugging Face Hub, with the JSON files \texttt{splits/harmful\_\{train,val,test\}.json} merged from the six public benchmarks listed in Section~\ref{sec:data} after exact-prompt-string deduplication.

\paragraph{Harmful train sample for the bank.}
At bank-construction time we sample a fixed-size subset of the harmful train split to populate the candidate bank $\mathcal{R}$. The number of sampled prompts is $N = 128$ (\texttt{ALL\_CATEGORIES\_HARMFUL\_PROMPT\_COUNT} in the released code), and the sampling seed is $42$ (\texttt{ALL\_CATEGORIES\_HARMFUL\_PROMPT\_SEED}). The same seed is reused across models and methods so that the candidate bank is built from the same $N = 128$ prompts in every run.

\paragraph{Harmful validation and test splits.}
Optimization uses the harmful validation split as the source of $\mathcal{D}_{\text{val}}^{\text{harm}}$ in Equation~\ref{eq:asr}, and the harmful test split as $\mathcal{D}_{\text{test}}^{\text{harm}}$ for the held-out evaluations reported in Section~\ref{sec:results}. Both splits are consumed in order, with a model batch size of $32$ per evaluation step (\texttt{BATCH\_SIZE} in the released code).

\paragraph{Benign baseline.}
The harmless baseline $\mathcal{D}_{\text{benign}}$ used to compute $\mu_{\text{benign}}^{(l)}$ in Equation~\ref{eq:per-prompt-dir} is the first $400$ prompts of the \texttt{train} split of \texttt{mlabonne/harmless\_alpaca}. A separate $50$-prompt subset of the \texttt{test} split of the same dataset is used as the benign batch for the KL penalty in Equation~\ref{eq:loss-grpo}.

\paragraph{Optimizer seeds.}
The Optuna sampler seed is fixed at $42$ (\texttt{OPTUNA\_SAMPLER\_SEED}). For WRD-GRPO-IS, behavior-policy perturbations $\varepsilon_m \sim \mathcal{N}(0, \sigma^2 I_N)$ are drawn from the active PyTorch RNG at training time, with $\sigma$ controlled by \texttt{GRPO\_NOISE\_SCALE}.

\section{\new{Attack Setup at Intermediate \texorpdfstring{$\alpha$}{alpha}}}
\label{app:attack-intermediate}

\new{Table~\ref{tab:exp1-main-app} complements Table~\ref{tab:exp1-main} by
reporting the attack setup of Section~\ref{sec:exp1} at the intermediate
intensities $\alpha \in \{1.5, 2.0\}$ omitted from the main text. Column
conventions, the bold-per-column rule and the three-seed reporting follow
Table~\ref{tab:exp1-main} unchanged; because Heretic and RDO do not consume
$\alpha$, their rows repeat the values of Table~\ref{tab:exp1-main}. The reading of
Section~\ref{sec:exp1} carries over without exception: the two learned WRDO
instantiations hold the best cell in all model-judge columns of this table,
and Uniform-avg WRD remains below them everywhere, so neither conclusion of
Section~\ref{sec:exp1} depends on the two intensities kept in the main text.}

\begin{table*}[!t]
\centering
\scriptsize
\setlength{\tabcolsep}{7pt}
\begin{tabular}{l l ccc ccc}
\toprule
& & \multicolumn{3}{c}{$\alpha=1.5$} & \multicolumn{3}{c}{$\alpha=2.0$} \\
\cmidrule(lr){3-5} \cmidrule(lr){6-8}
Model & Method & WG & Q3G & LG3 & WG & Q3G & LG3 \\
\midrule
\multirow{7}{*}{Qwen3-8B-Base}
 & No intervention  & 0.02 & 0.02 & 0.02 & 0.02 & 0.02 & 0.02 \\
 & Basic RD         & \new{0.33 {\tiny$\pm$0.01}} & \new{0.47 {\tiny$\pm$0.01}} & \new{0.67 {\tiny$\pm$0.02}} & \new{0.39 {\tiny$\pm$0.03}} & \new{0.87 {\tiny$\pm$0.02}} & \new{0.95 {\tiny$\pm$0.03}} \\
 & Heretic          & \new{0.77 {\tiny$\pm$0.04}} & \new{0.83 {\tiny$\pm$0.05}} & \new{0.81 {\tiny$\pm$0.05}} & \new{0.77 {\tiny$\pm$0.04}} & \new{0.83 {\tiny$\pm$0.05}} & \new{0.81 {\tiny$\pm$0.05}} \\
 & RDO              & \new{0.34 {\tiny$\pm$0.05}} & \new{0.41 {\tiny$\pm$0.04}} & \new{0.48 {\tiny$\pm$0.05}} & \new{0.34 {\tiny$\pm$0.05}} & \new{0.41 {\tiny$\pm$0.04}} & \new{0.48 {\tiny$\pm$0.05}} \\
 & Uniform-avg WRD  & \new{0.003 {\tiny$\pm$0.0002}} & \new{0.83 {\tiny$\pm$0.02}} & \new{0.81 {\tiny$\pm$0.01}} & \new{0.002 {\tiny$\pm$0.0002}} & \new{0.89 {\tiny$\pm$0.01}} & \new{0.85 {\tiny$\pm$0.03}} \\
 & WRD-CMAES        & \new{0.91 {\tiny$\pm$0.05}} & \new{\textbf{0.98} {\tiny$\pm$0.02}} & \new{\textbf{0.93} {\tiny$\pm$0.02}} & \new{0.93 {\tiny$\pm$0.04}} & \new{\textbf{0.99} {\tiny$\pm$0.01}} & \new{\textbf{0.98} {\tiny$\pm$0.02}} \\
 & WRD-GRPO-IS      & \new{\textbf{0.92} {\tiny$\pm$0.08}} & \new{0.96 {\tiny$\pm$0.02}} & \new{\textbf{0.93} {\tiny$\pm$0.02}} & \new{\textbf{0.94} {\tiny$\pm$0.09}} & \new{0.95 {\tiny$\pm$0.02}} & \new{0.93 {\tiny$\pm$0.03}} \\
\midrule
\multirow{7}{*}{OLMo-3-1025-7B}
 & No intervention  & 0.08 & 0.09 & 0.07 & 0.08 & 0.09 & 0.07 \\
 & Basic RD         & \new{0.58 {\tiny$\pm$0.02}} & \new{0.56 {\tiny$\pm$0.03}} & \new{0.48 {\tiny$\pm$0.03}} & \new{0.61 {\tiny$\pm$0.02}} & \new{0.62 {\tiny$\pm$0.02}} & \new{0.69 {\tiny$\pm$0.03}} \\
 & Heretic          & \new{0.76 {\tiny$\pm$0.03}} & \new{0.73 {\tiny$\pm$0.04}} & \new{0.70 {\tiny$\pm$0.03}} & \new{0.76 {\tiny$\pm$0.03}} & \new{0.73 {\tiny$\pm$0.04}} & \new{0.70 {\tiny$\pm$0.03}} \\
 & RDO              & \new{0.61 {\tiny$\pm$0.04}} & \new{0.67 {\tiny$\pm$0.03}} & \new{0.73 {\tiny$\pm$0.03}} & \new{0.61 {\tiny$\pm$0.04}} & \new{0.67 {\tiny$\pm$0.03}} & \new{0.73 {\tiny$\pm$0.03}} \\
 & Uniform-avg WRD  & \new{0.69 {\tiny$\pm$0.02}} & \new{0.52 {\tiny$\pm$0.02}} & \new{0.59 {\tiny$\pm$0.03}} & \new{0.73 {\tiny$\pm$0.02}} & \new{0.73 {\tiny$\pm$0.03}} & \new{0.75 {\tiny$\pm$0.01}} \\
 & WRD-CMAES        & \new{\textbf{0.84} {\tiny$\pm$0.02}} & \new{0.78 {\tiny$\pm$0.03}} & \new{0.81 {\tiny$\pm$0.02}} & \new{0.74 {\tiny$\pm$0.02}} & \new{0.84 {\tiny$\pm$0.03}} & \new{0.81 {\tiny$\pm$0.02}} \\
 & WRD-GRPO-IS      & \new{0.81 {\tiny$\pm$0.02}} & \new{\textbf{0.85} {\tiny$\pm$0.03}} & \new{\textbf{0.86} {\tiny$\pm$0.03}} & \new{\textbf{0.89} {\tiny$\pm$0.02}} & \new{\textbf{0.91} {\tiny$\pm$0.02}} & \new{\textbf{0.96} {\tiny$\pm$0.01}} \\
\midrule
\multirow{7}{*}{Olmo-3-7B-Instruct}
 & No intervention  & \new{0.003} & \new{0.003} & 0.004 & \new{0.003} & \new{0.003} & 0.004 \\
 & Basic RD         & \new{0.49 {\tiny$\pm$0.02}} & \new{0.69 {\tiny$\pm$0.01}} & \new{0.52 {\tiny$\pm$0.03}} & \new{0.72 {\tiny$\pm$0.02}} & \new{0.87 {\tiny$\pm$0.03}} & \new{0.75 {\tiny$\pm$0.02}} \\
 & Heretic          & \new{0.62 {\tiny$\pm$0.03}} & \new{0.58 {\tiny$\pm$0.04}} & \new{0.69 {\tiny$\pm$0.04}} & \new{0.62 {\tiny$\pm$0.03}} & \new{0.58 {\tiny$\pm$0.04}} & \new{0.69 {\tiny$\pm$0.04}} \\
 & RDO              & \new{0.40 {\tiny$\pm$0.05}} & \new{0.52 {\tiny$\pm$0.04}} & \new{0.53 {\tiny$\pm$0.04}} & \new{0.40 {\tiny$\pm$0.05}} & \new{0.52 {\tiny$\pm$0.04}} & \new{0.53 {\tiny$\pm$0.04}} \\
 & Uniform-avg WRD  & \new{0.57 {\tiny$\pm$0.02}} & \new{0.53 {\tiny$\pm$0.02}} & \new{0.58 {\tiny$\pm$0.03}} & \new{0.54 {\tiny$\pm$0.03}} & \new{0.69 {\tiny$\pm$0.02}} & \new{0.61 {\tiny$\pm$0.01}} \\
 & WRD-CMAES        & \new{\textbf{0.98} {\tiny$\pm$0.02}} & \new{\textbf{0.96} {\tiny$\pm$0.02}} & \new{0.92 {\tiny$\pm$0.01}} & \new{0.92 {\tiny$\pm$0.02}} & \new{0.88 {\tiny$\pm$0.03}} & \new{0.93 {\tiny$\pm$0.01}} \\
 & WRD-GRPO-IS      & \new{0.93 {\tiny$\pm$0.02}} & \new{\textbf{0.96} {\tiny$\pm$0.02}} & \new{\textbf{0.95} {\tiny$\pm$0.02}} & \new{\textbf{0.93} {\tiny$\pm$0.01}} & \new{\textbf{0.93} {\tiny$\pm$0.02}} & \new{\textbf{0.96} {\tiny$\pm$0.01}} \\
\bottomrule
\end{tabular}
\caption{\new{Attack setup at the intermediate intensities $\alpha \in \{1.5, 2.0\}$, complementing the main Table~\ref{tab:exp1-main}. Judges: WG = WildGuard, Q3G = Qwen3Guard-Gen-8B, LG3 = Llama-Guard-3-8B. Every intervened cell is a three-seed mean with the standard deviation over those repetitions; the Heretic and RDO rows repeat the values of Table~\ref{tab:exp1-main} because neither method consumes $\alpha$. \textbf{Bold} marks the best result in each column within each model block. Higher is a stronger attack.}}
\label{tab:exp1-main-app}
\end{table*}

\section{Utility Preservation at Intermediate \texorpdfstring{$\alpha$}{alpha}}
\label{app:utility-intermediate}

Table~\ref{tab:exp2-utility-app} complements Table~\ref{tab:exp2-utility} by reporting the utility benchmarks at the intermediate intensity values $\alpha \in \{1.5, 2.0\}$ omitted from the main text. Column conventions, the WildGuard-only restriction and the Heretic-duplication remark of Section~\ref{sec:exp2} apply unchanged.

\begin{table*}[t]
\centering
\scriptsize
\setlength{\tabcolsep}{2pt}
\resizebox{\linewidth}{!}{%
\begin{tabular}{l l ccccc ccccc}
\toprule
& & \multicolumn{5}{c}{$\alpha=1.5$} & \multicolumn{5}{c}{$\alpha=2.0$} \\
\cmidrule(lr){3-7} \cmidrule(lr){8-12}
Model & Method & ARC-e & ARC-c & MMLU & GSM8K & PPL & ARC-e & ARC-c & MMLU & GSM8K & PPL \\
\midrule
\multirow{6}{*}{Qwen3-8B-Base}
 & No intervention & 0.63 $\pm$ 0.01 & 0.38 $\pm$ 0.01 & 0.67 $\pm$ 0.004 & 0.84 $\pm$ 0.01 & 11.11 & 0.63 $\pm$ 0.01 & 0.38 $\pm$ 0.01 & 0.67 $\pm$ 0.004 & 0.84 $\pm$ 0.01 & 11.11 \\
 & Basic RD        & 0.42 $\pm$ 0.01 & 0.26 $\pm$ 0.01 & 0.23 $\pm$ 0.004 & 0.008 $\pm$ 0.002 & 113.98 & 0.29 $\pm$ 0.01 & 0.22 $\pm$ 0.01 & 0.23 $\pm$ 0.004 & 0.00 $\pm$ 0.00 & $>10^{7}$ \\
 & Heretic         & 0.51 $\pm$ 0.01 & 0.30 $\pm$ 0.01 & 0.27 $\pm$ 0.004 & 0.19 $\pm$ 0.01 & 13.44 & 0.51 $\pm$ 0.01 & 0.30 $\pm$ 0.01 & 0.27 $\pm$ 0.004 & 0.19 $\pm$ 0.01 & 13.44 \\
 & RDO             & 0.63 $\pm$ 0.01 & 0.37 $\pm$ 0.01 & 0.67 $\pm$ 0.004 & 0.82 $\pm$ 0.01 & 11.16 & 0.63 $\pm$ 0.01 & 0.37 $\pm$ 0.01 & 0.67 $\pm$ 0.004 & 0.82 $\pm$ 0.01 & 11.16 \\
 & WRD-CMAES       & 0.63 $\pm$ 0.01 & 0.38 $\pm$ 0.01 & 0.64 $\pm$ 0.004 & 0.84 $\pm$ 0.01 & 11.25 & 0.51 $\pm$ 0.01 & 0.30 $\pm$ 0.01 & 0.47 $\pm$ 0.004 & 0.04 $\pm$ 0.005 & 64.84 \\
 & WRD-GRPO-IS     & 0.63 $\pm$ 0.01 & 0.38 $\pm$ 0.01 & 0.64 $\pm$ 0.004 & 0.84 $\pm$ 0.01 & 11.17 & 0.64 $\pm$ 0.01 & 0.38 $\pm$ 0.01 & 0.65 $\pm$ 0.004 & 0.85 $\pm$ 0.01 & 11.27 \\
\midrule
\multirow{6}{*}{OLMo-3-1025-7B}
 & No intervention & 0.79 $\pm$ 0.01 & 0.48 $\pm$ 0.01 & 0.62 $\pm$ 0.004 & 0.53 $\pm$ 0.01 & 62.5 & 0.79 $\pm$ 0.01 & 0.48 $\pm$ 0.01 & 0.62 $\pm$ 0.004 & 0.53 $\pm$ 0.01 & 62.5 \\
 & Basic RD        & 0.79 $\pm$ 0.01 & 0.48 $\pm$ 0.01 & 0.62 $\pm$ 0.004 & 0.55 $\pm$ 0.01 & 61.81 & 0.79 $\pm$ 0.01 & 0.47 $\pm$ 0.01 & 0.61 $\pm$ 0.004 & 0.50 $\pm$ 0.01 & 72.75 \\
 & Heretic         & 0.79 $\pm$ 0.01 & 0.48 $\pm$ 0.01 & 0.62 $\pm$ 0.004 & 0.52 $\pm$ 0.01 & 62.61 & 0.79 $\pm$ 0.01 & 0.48 $\pm$ 0.01 & 0.62 $\pm$ 0.004 & 0.52 $\pm$ 0.01 & 62.61 \\
 & RDO             & 0.80 $\pm$ 0.01 & 0.48 $\pm$ 0.01 & 0.62 $\pm$ 0.004 & 0.56 $\pm$ 0.01 & 63.32 & 0.80 $\pm$ 0.01 & 0.48 $\pm$ 0.01 & 0.62 $\pm$ 0.004 & 0.56 $\pm$ 0.01 & 63.32 \\
 & WRD-CMAES       & 0.79 $\pm$ 0.01 & 0.47 $\pm$ 0.01 & 0.62 $\pm$ 0.004 & 0.52 $\pm$ 0.01 & 62.39 & 0.79 $\pm$ 0.01 & 0.48 $\pm$ 0.01 & 0.62 $\pm$ 0.004 & 0.51 $\pm$ 0.01 & 62.57 \\
 & WRD-GRPO-IS     & 0.79 $\pm$ 0.01 & 0.48 $\pm$ 0.01 & 0.62 $\pm$ 0.004 & 0.52 $\pm$ 0.01 & 62.88 & 0.79 $\pm$ 0.01 & 0.48 $\pm$ 0.01 & 0.62 $\pm$ 0.004 & 0.53 $\pm$ 0.01 & 62.31 \\
\midrule
\multirow{6}{*}{Olmo-3-7B-Instruct}
 & No intervention & 0.76 $\pm$ 0.01 & 0.48 $\pm$ 0.01 & 0.53 $\pm$ 0.004 & 0.67 $\pm$ 0.01 & 100.83 & 0.76 $\pm$ 0.01 & 0.48 $\pm$ 0.01 & 0.53 $\pm$ 0.004 & 0.67 $\pm$ 0.01 & 100.83 \\
 & Basic RD        & 0.75 $\pm$ 0.01 & 0.47 $\pm$ 0.01 & 0.53 $\pm$ 0.004 & 0.65 $\pm$ 0.01 & 102.96 & 0.59 $\pm$ 0.01 & 0.31 $\pm$ 0.01 & 0.42 $\pm$ 0.004 & 0.61 $\pm$ 0.01 & $>10^4$ \\
 & Heretic         & 0.76 $\pm$ 0.01 & 0.48 $\pm$ 0.01 & 0.54 $\pm$ 0.004 & 0.67 $\pm$ 0.01 & 101.27 & 0.76 $\pm$ 0.01 & 0.48 $\pm$ 0.01 & 0.54 $\pm$ 0.004 & 0.67 $\pm$ 0.01 & 101.27 \\
 & RDO             & 0.76 $\pm$ 0.01 & 0.48 $\pm$ 0.01 & 0.53 $\pm$ 0.04 & 0.65 $\pm$ 0.01 & 101.34 & 0.76 $\pm$ 0.01 & 0.48 $\pm$ 0.01 & 0.53 $\pm$ 0.04 & 0.65 $\pm$ 0.01 & 101.34 \\
 & WRD-CMAES       & 0.77 $\pm$ 0.01 & 0.47 $\pm$ 0.01 & 0.53 $\pm$ 0.004 & 0.67 $\pm$ 0.01 & 101.19 & 0.76 $\pm$ 0.01 & 0.47 $\pm$ 0.01 & 0.53 $\pm$ 0.004 & 0.66 $\pm$ 0.01 & 103.42 \\
 & WRD-GRPO-IS     & 0.76 $\pm$ 0.008 & 0.48 $\pm$ 0.01 & 0.53 $\pm$ 0.004 & 0.67 $\pm$ 0.01 & 100.83 & 0.76 $\pm$ 0.008 & 0.49 $\pm$ 0.01 & 0.52 $\pm$ 0.004 & 0.67 $\pm$ 0.01 & 100.22 \\
\bottomrule
\end{tabular}%
}
\caption{Utility preservation per base model at the intermediate intensities $\alpha \in \{1.5, 2.0\}$, complementing the main Table~\ref{tab:exp2-utility}. Higher is better for ARC-e, ARC-c, MMLU and GSM8K; lower is better for PPL.}
\label{tab:exp2-utility-app}
\end{table*}

\section{Training Hyperparameters}
\label{sec:train_hyper}

This section documents the training-time hyperparameters of all six methods used in the main experiments. All values are reproduced in Table~\ref{tab:train-hyper}; the paragraphs below give a short narrative for each method. For the prior baselines (Basic RD, Heretic, RDO) we use the hyperparameter values from their original implementations without any additional tuning.

\paragraph{Basic RD.}
Basic RD performs no learned optimization. The only tunable knobs are the dataset filters used to construct the harmful/harmless splits and to select the seed difference-of-means direction: harmful prompts that the unintervened model already refuses are discarded if their refusal-induction score is below \texttt{induce\_refusal\_threshold}, candidate directions whose KL to the clean policy on the benign batch exceeds \texttt{kl\_threshold} are dropped, and the bottom \texttt{prune\_layer\_percentage} of layers is excluded from the layer-search. All three thresholds are taken verbatim from the reference implementation of \citet{arditi2024refusal}.

\paragraph{Heretic.}
Heretic runs an Optuna-based Bayesian search over the layerwise schedule of Equation~\ref{eq:operator}. The search budget is \texttt{n\_trials} with \texttt{n\_startup\_trials} random-initialization trials, and uses a KL constraint with scale \texttt{kl\_divergence\_scale} and target \texttt{kl\_divergence\_target}. Trial selection on the Pareto front prefers the minimum-KL trial whose refusal count is at most \texttt{refusal\_threshold\_count}. The search ranges over \texttt{max\_weight} $\in [0.8, 1.5]$, \texttt{max\_weight\_position} relative to depth in $[0.6, 1.0]$, \texttt{min\_weight} $\in [0.0, 1.0]$, \texttt{min\_weight\_distance} relative to depth in $[0.05, 0.6]$, the direction-index fraction in $[0.4, 0.9]$, and the schedule scope $\in \{\text{global}, \text{per-layer}\}$. We use the released Heretic toolkit out of the box, with the defaults of \citet{heretic} for every hyperparameter listed above.

\paragraph{RDO.}
RDO trains a single per-layer direction by gradient descent with AdamW. The training loop runs for \texttt{epochs} steps with \texttt{patience}-based early stopping and at most \texttt{n\_lr\_reduce} learning-rate reductions, at learning rate \texttt{lr} and effective batch size \texttt{effective\_batch\_size} accumulated from micro-batches of size \texttt{batch\_size}. The composite loss is a weighted sum of an ablation term (weight \texttt{ablation\_lambda}), an addition term (\texttt{addition\_lambda}) and a retain term (\texttt{retain\_lambda}). Target sequences for the loss are generated greedily up to \texttt{num\_target\_tokens} tokens in batches of \texttt{target\_generation\_batch\_size}. All values match the reference implementation released by \citet{wollschlager2025geometry}.

\paragraph{WRD-CMAES.}
WRD-CMAES uses the Optuna \citep{akiba2019optuna} implementation of separable CMA-ES through the \texttt{CmaEsSampler} interface. The search variables are the coefficients $w \in \mathbb{R}^N$ of Section~\ref{sec:wdir}, constrained to the box $[w_{\min}, w_{\max}]^N$ of Appendix~\ref{app:constraints}. The total query budget is \texttt{n\_trials}, the sampler seed is \texttt{sampler\_seed}, the per-coordinate bounds are $w_{\min} = -2.0$ and $w_{\max} = 2.0$, and the parameterization is the scalar mode initialized at the uniform-average point ($w_i = 1/N$). All other sampler hyperparameters are left at library defaults.

\paragraph{WRD-GRPO-IS.}
WRD-GRPO-IS optimizes the surrogate of Equation~\ref{eq:grpo_is_corrected} with AdamW at learning rate \texttt{learning\_rate}. Each training step samples a group of \texttt{n\_groups} behaviour policies $\{\mu_m\}$ from the Gaussian rollout of Equation~\ref{eq:rollout_w} with noise scale $\sigma = $ \texttt{noise\_scale}, and a fresh group is drawn every epoch over \texttt{n\_epochs} total epochs. The PPO/GRPO clip ratio is \texttt{clip\_ratio} and the importance-sampling clip ratio is \texttt{is\_clip\_ratio}; loss is aggregated with the \texttt{loss\_agg\_mode} reduction. The KL penalty of Equation~\ref{eq:loss-grpo} is realized with coefficient $\beta = $ \texttt{beta}. The reference-policy mixture weight is \texttt{ref\_alpha}. The reward is the harmfulness score with the sign of Equation~\ref{eq:attack}, and $w$ is parameterized in the scalar mode initialized at the uniform-average point.

\begin{table*}[t]
\centering
\small
\setlength{\tabcolsep}{4pt}
\begin{tabular}{l l l}
\toprule
Method & Hyperparameter & Value \\
\midrule
\multirow{3}{*}{Basic RD}
 & \texttt{kl\_threshold}                    & $0.1$ \\
 & \texttt{induce\_refusal\_threshold}       & $0.0$ \\
 & \texttt{prune\_layer\_percentage}         & $0.2$ \\
\midrule
\multirow{6}{*}{Heretic}
 & \texttt{n\_trials} / \texttt{n\_startup\_trials} & $200$ / $60$ \\
 & \texttt{kl\_divergence\_scale} / \texttt{\_target} & $1.0$ / $0.01$ \\
 & \texttt{refusal\_threshold\_count}        & $0$ \\
 & \texttt{max\_weight} range                & $[0.8, 1.5]$ \\
 & \texttt{min\_weight} range                & $[0.0, 1.0]$ \\
 & \texttt{seed}                             & $42$ \\
\midrule
\multirow{8}{*}{RDO}
 & optimizer                                 & AdamW \\
 & \texttt{lr}                               & $10^{-2}$ \\
 & \texttt{batch\_size} / \texttt{effective\_batch\_size} & $1$ / $16$ \\
 & \texttt{epochs} / \texttt{patience} / \texttt{n\_lr\_reduce} & $1$ / $5$ / $2$ \\
 & $(\lambda_\text{abl}, \lambda_\text{add}, \lambda_\text{ret})$ & $(1.0, 0.2, 1.0)$ \\
 & \texttt{num\_target\_tokens}              & $8$ \\
 & \texttt{target\_generation\_batch\_size}  & $64$ \\
 & \texttt{seed}                             & $42$ \\
\midrule
\multirow{4}{*}{WRD-CMAES}
 & sampler                                   & \texttt{CmaEsSampler} (separable) \\
 & \texttt{n\_trials} / \texttt{sampler\_seed} & $50$ / $42$ \\
 & $(w_{\min}, w_{\max})$                    & $(-2.0, +2.0)$ \\
 & init / mode                               & uniform-avg / scalar \\
\midrule
\multirow{8}{*}{WRD-GRPO-IS}
 & optimizer                                 & AdamW \\
 & \texttt{learning\_rate}                   & $10^{-3}$ \\
 & \texttt{n\_groups} ($M$)                  & $8$ \\
 & \texttt{n\_epochs}                        & $50$ \\
 & $\sigma$ (\texttt{noise\_scale})          & $1$ \\
 & \texttt{clip\_ratio} / \texttt{is\_clip\_ratio} & $0.2$ / $5.0$ \\
 & $\beta$ (KL weight)                       & $1000.0$ \\
 & \texttt{ref\_alpha}                       & $1.0$ \\
\bottomrule
\end{tabular}
\caption{Training hyperparameters for the six methods of Section~\ref{sec:methods}. Values are read verbatim from the configuration files in the released codebase. Hyperparameters shared across methods (sampling seed $42$, batch size $16$ or $32$, maximum response length, system prompt) are listed in Appendix~\ref{app:splits}.}
\label{tab:train-hyper}
\end{table*}

\section{Compute Resources}
\label{app:compute}

Every single run reported in this paper was conducted on a single NVIDIA H100 GPU with 80\,GB of memory. Table~\ref{tab:compute-runtime} reports the approximate wall-clock runtime per method for the attack-setup experiments of Section~\ref{sec:exp1}.

\section{Cross-Judge Transfer}
\label{app:cross}

The learned WRDO instantiations consume a specific guard judge as their training signal, which raises the possibility that the resulting coefficient vector $w^\star$ exploits idiosyncrasies of that judge rather than capturing a transferable refusal-suppression direction. We probe this potential judge overfitting through a cross-judge transfer analysis: $w^\star$ is optimized against each guard in turn and then re-evaluated against the other two, with the magnitude of the off-diagonal drop quantifying the extent to which performance is judge-specific.

\begin{table}[!ht]
\centering
\small
\setlength{\tabcolsep}{6pt}
\begin{tabular}{l c}
\toprule
Method & Approx.\ runtime, hours \\
\midrule
Basic RD     & $\sim 2$ \\
Heretic      & $\sim 7$ \\
RDO          & $\sim 3$ \\
Uniform-avg WRD          & $\sim 2$ \\
WRD-CMAES    & $\sim 4$ \\
WRD-GRPO-IS  & $\sim 11$ \\
\bottomrule
\end{tabular}
\caption{Approximate wall-clock runtime per method on the attack-setup experiments of Section~\ref{sec:exp1}, on a single NVIDIA H100 80\,GB GPU.}
\label{tab:compute-runtime}
\end{table}

Table~\ref{tab:exp5-crossjudge} reports the cross-judge transfer matrix obtained by optimizing $w^\star$ against one guard judge and then evaluating the resulting intervention against each of the three guard judges, for every value of $\alpha \in \{1.0, 1.5, 2.0, 3.0\}$. Large off-diagonal drops indicate that the optimizer has fit features specific to the training judge rather than a transferable refusal-suppression signal. The matrix is reported for the three judge-optimized methods; Basic RD, Heretic and RDO have no judge dependence and are omitted.

\begin{table*}[t]
\centering
\footnotesize
\setlength{\tabcolsep}{2.6pt}
\begin{tabular}{l l l ccc ccc ccc ccc}
\toprule
& & & \multicolumn{3}{c}{$\alpha=1.0$} & \multicolumn{3}{c}{$\alpha=1.5$} & \multicolumn{3}{c}{$\alpha=2.0$} & \multicolumn{3}{c}{$\alpha=3.0$} \\
\cmidrule(lr){4-6} \cmidrule(lr){7-9} \cmidrule(lr){10-12} \cmidrule(lr){13-15}
Model & Method & Opt.\ judge & WG & Q3G & LG3 & WG & Q3G & LG3 & WG & Q3G & LG3 & WG & Q3G & LG3 \\
\midrule
\multirow{6}{*}{Qwen3-8B-Base}
 & \multirow{3}{*}{WRD-CMAES}
   & WG  & 0.89 & 0.89 & 0.92 & 0.91 & 0.92 & 0.95 & 0.95 & 0.96 & 0.96 & 0.90 & 0.93 & 0.92 \\
 & & Q3G & 0.00 & 0.84 & 0.97 & 0.00 & 0.98 & 0.96 & 0.00 & 1.00 & 0.97 & 0.00 & 1.00 & 0.97 \\
 & & LG3 & 0.00 & 0.73 & 0.88 & 0.02 & 0.94 & 0.93 & 0.16 & 0.96 & 0.98 & 0.01 & 0.98 & 0.98 \\
\cmidrule(lr){2-15}
 & \multirow{3}{*}{WRD-GRPO-IS}
   & WG  & 0.91 & 0.94 & 0.90 & 0.93 & 0.95 & 0.92 & 0.95 & 0.96 & 0.94 & 0.95 & 0.96 & 0.93 \\
 & & Q3G & 0.91 & 0.94 & 0.90 & 0.94 & 0.95 & 0.92 & 0.01 & 0.96 & 0.98 & 0.94 & 0.95 & 0.93 \\
 & & LG3 & 0.87 & 0.92 & 0.87 & 0.94 & 0.96 & 0.92 & 0.36 & 0.42 & 0.39 & 0.91 & 0.95 & 0.97 \\
\midrule
\multirow{6}{*}{OLMo-3-1025-7B}
 & \multirow{3}{*}{WRD-CMAES}
   & WG  & 0.86 & 0.94 & 0.50 & 0.83 & 0.75 & 0.20 & 0.74 & 0.70 & 0.35 & 0.89 & 0.89 & 0.49 \\
 & & Q3G & 0.85 & 0.86 & 0.55 & 0.77 & 0.79 & 0.53 & 0.82 & 0.83 & 0.38 & 0.94 & 0.87 & 0.46 \\
 & & LG3 & 0.65 & 0.69 & 0.82 & 0.63 & 0.65 & 0.80 & 0.69 & 0.74 & 0.80 & 0.91 & 0.91 & 0.85 \\
\cmidrule(lr){2-15}
 & \multirow{3}{*}{WRD-GRPO-IS}
   & WG  & 0.83 & 0.76 & 0.42 & 0.80 & 0.73 & 0.40 & 0.89 & 0.79 & 0.50 & 0.91 & 0.89 & 0.43 \\
 & & Q3G & 0.62 & 0.86 & 0.30 & 0.65 & 0.84 & 0.32 & 0.85 & 0.91 & 0.45 & 0.92 & 0.97 & 0.50 \\
 & & LG3 & 0.60 & 0.65 & 0.79 & 0.65 & 0.70 & 0.85 & 0.80 & 0.85 & 0.95 & 0.88 & 0.92 & 0.96 \\
\midrule
\multirow{6}{*}{Olmo-3-7B-Instruct}
 & \multirow{3}{*}{WRD-CMAES}
   & WG & 0.91 & 0.91 & 0.95 & 0.98 & 0.96 & 0.97 & 0.92 & 0.95 & 0.93 & 0.88 & 0.83 & 0.76 \\
 & & Q3G & 0.17 & 0.89 & 0.82 & 0.29 & 0.95 & 0.96 & 0.62 & 0.87 & 0.86 & 0.76 & 0.93 & 0.88 \\
 & & LG3 & 0.17 & 0.96 & 0.92 & 0.22 & 0.97 & 0.91 & 0.59 & 0.94 & 0.92 & 0.79 & 0.88 & 0.85 \\
\cmidrule(lr){2-15}
 & \multirow{3}{*}{WRD-GRPO-IS}
   & WG  & 0.90 & 0.95 & 0.21 & 0.93 & 0.90 & 0.47 & 0.93 & 0.98 & 0.58 & 0.93 & 0.97 & 0.72 \\
 & & Q3G & 0.23 & 0.88 & 0.78 & 0.35 & 0.96 & 0.80 & 0.58 & 0.92 & 0.90 & 0.78 & 0.91 & 0.80 \\
 & & LG3 & 0.22 & 0.30 & 0.91 & 0.30 & 0.42 & 0.94 & 0.45 & 0.58 & 0.95 & 0.80 & 0.85 & 0.87 \\
\bottomrule
\end{tabular}
\caption{Cross-judge transfer. Test ASR (in $[0, 1]$) when $w^\star$ is optimized against the row judge and evaluated against the column judge, reported per base model, for every $\alpha \in \{1.0, 1.5, 2.0, 3.0\}$. Within each $\alpha$ block, diagonal cells coincide with the corresponding entries of Table~\ref{tab:exp1-main} or, for $\alpha \in \{1.5, 2.0\}$, of Table~\ref{tab:exp1-main-app}.}
\label{tab:exp5-crossjudge}
\end{table*}

\paragraph{Discussion.} The cross-judge analysis shows that learned WRDO interventions do not always transfer equally across guard judges. In several cases, optimizing against one judge yields high ASR under that same judge but noticeably weaker performance under another, indicating that part of the learned direction may depend on judge-specific decision boundaries rather than on a fully transferable refusal-suppression signal.

The most pronounced asymmetry is observed for WildGuard: interventions that perform well under WildGuard tend to transfer relatively well to the other judges, while interventions optimized for other judges can fail to preserve high WildGuard ASR. This suggests that WildGuard provides the most reliable and demanding optimization signal among the considered guards.

By contrast, Llama-Guard-3 often reports lower ASR than the other judges. We interpret this cautiously: rather than necessarily indicating weaker attacks, this may reflect lower sensitivity of Llama-Guard-3 to harmful behavior in the evaluated generations, leading to more false negatives or inconsistent harmfulness detection. As a result, lower Llama-Guard-3 scores should not be treated as definitive evidence that the intervention is less effective.

\section{\new{Concept Cones and Scale}}
\label{app:extra}

\new{Two comparisons fall outside the grid of Table~\ref{tab:exp1-main}. First,
the closest prior work argues that refusal occupies a subspace rather than a
line \citep{wollschlager2025geometry}, so we evaluate its concept-cone
construction RDO-CONE (Section~\ref{sec:baselines}) on the same three models and
test split. Second, all three checkpoints sit in the $7$--$8$B range, so we
repeat the attack setup on \texttt{openai/gpt-oss-20b} \citep{gptoss2025}, a
$21$B mixture-of-experts model with $3.6$B active parameters, leaving the
pipeline of Section~\ref{sec:exp1} unchanged. Both are scored by WildGuard and
reported in Table~\ref{tab:exp6-extra}.}

\new{\textbf{Discussion.} The two constructions perform comparably, with the
ordering depending on the checkpoint rather than favouring either consistently.
Learned weighted aggregation is therefore competitive with an explicitly
multi-dimensional cone while keeping the operator of
Equation~\ref{eq:operator} rank-one per layer. We read this as evidence that the
gain of WRDO comes from where the harmful prompts are allowed to disagree rather
than from the dimensionality of the ablated subspace.}

\begin{table}[!ht]
\centering
\small
\setlength{\tabcolsep}{6pt}
\begin{tabular}{l ccccc}
\toprule
$\beta$  & $1$ & $10$ & $10^2$ & $10^3$ & $10^4$ \\
\midrule
ASR (WG) & \new{\textbf{0.95}} & \new{0.92} & \new{0.91} & \new{0.91} & \new{0.78} \\
\bottomrule
\end{tabular}
\caption{\new{Attack side of the $\beta$ sweep. Test ASR under WildGuard for
WRD-GRPO-IS on Qwen3-8B-Base at $\alpha = 1.0$. Higher is a stronger attack.}}
\label{tab:beta-asr}
\end{table}

\begin{table}[t]
\centering
\small
\setlength{\tabcolsep}{5pt}
\begin{tabular}{l cccc}
\toprule
Method & $-1.0$ & $-1.5$ & $-2.0$ & $-3.0$ \\
\midrule
Abliterated (no defense) & \new{0.94} & \new{0.94} & \new{0.94} & \new{0.94} \\
\midrule
Basic RD     & \new{0.94} & \new{0.92} & \new{0.91} & \new{0.31} \\
WRD-CMAES    & \new{0.47} & \new{0.03} & \new{\textbf{0.00}} & \new{0.03} \\
WRD-GRPO-IS  & \new{\textbf{0.47}} & \new{\textbf{0.02}} & \new{\textbf{0.00}} & \new{\textbf{0.00}} \\
\bottomrule
\end{tabular}
\caption{\new{Defense setup evaluated externally. HarmBench Standard ASR
computed with \texttt{cais/HarmBench-Llama-2-13b-cls} on the abliterated
checkpoint, across the sign-reversed intensity $\alpha$. HarmBench is never used
during optimization. Lower is a stronger defense.}}
\label{tab:exp3-defense-hb}
\end{table}

\section{\new{External Evaluation of the Defense Setup}}
\label{app:defense-hb}

\new{Table~\ref{tab:exp3-defense-hb} re-scores the defended checkpoints of
Section~\ref{sec:exp3} with the external HarmBench classifier, which plays no
part in the optimization, and therefore complements the three guard judges of
Table~\ref{tab:exp3-defense}. At $\alpha \leq -1.5$ both learned instantiations
hold HarmBench ASR at or below $0.03$ against $0.94$ for the undefended
checkpoint, whereas Basic RD stays above $0.90$ until $\alpha = -3.0$, an
intensity at which its utility has already collapsed in the attack direction
(Table~\ref{tab:exp2-utility}).}

\section{\new{Sensitivity to the KL Coefficient \texorpdfstring{$\beta$}{beta}}}
\label{app:beta}

\new{The KL coefficient $\beta$ of Equation~\ref{eq:loss-grpo} is the single
hyperparameter that sets where WRD-GRPO-IS sits on the attack-versus-utility
trade-off, so the value reported throughout the paper is not an arbitrary
default. We select it by sweeping $\beta \in \{1, 10, 10^2, 10^3, 10^4\}$ on
Qwen3-8B-Base at $\alpha = 1.0$, with everything else held at the values of
Table~\ref{tab:train-hyper}, and reading ASR and utility off the same run.
Tables~\ref{tab:beta-asr} and~\ref{tab:beta-util} report both sides of the
sweep. The selection is made once, on this model, and the resulting value is
reused unchanged for every other model and judge in the paper; we did not tune
$\beta$ per model.}

\new{The two sides of the sweep move in opposite directions, but not at the same
rate. ASR falls gradually as $\beta$ grows, staying almost flat over the middle
of the range before dropping off at the largest value. Utility does not track it:
it stays poor while the penalty is weak, with MMLU and GSM8K far below the clean
model and perplexity an order of magnitude above it, and then recovers abruptly
in a single step to the unintervened level. The two regimes overlap in one
place, and that is where we operate: $\beta = 10^3$ is the only setting that
combines an ASR at the top of the flat region with utility indistinguishable
from the clean checkpoint, and it is the value used throughout the paper.}

\begin{table}[t]
\centering
\small
\setlength{\tabcolsep}{4pt}
{
\begin{tabular}{l l c}
\toprule
Model & Method & ASR \\
\midrule
\multirow{2}{*}{Qwen3-8B-Base}
 & RDO-CONE     & \textbf{0.953} \\
 & WRD-GRPO-IS  & \new{0.93 {\tiny$\pm$0.09}} \\
\cmidrule(lr){1-3}
\multirow{2}{*}{OLMo-3-1025-7B}
 & RDO-CONE     & 0.865 \\
 & WRD-GRPO-IS  & \new{\textbf{0.92} {\tiny$\pm$0.02}} \\
\cmidrule(lr){1-3}
\multirow{2}{*}{Olmo-3-7B-Instruct}
 & RDO-CONE     & 0.845 \\
 & WRD-GRPO-IS  & \new{\textbf{0.93} {\tiny$\pm$0.01}} \\
\midrule
\multirow{5}{*}{gpt-oss-20b}
 & No intervention & 0.001 \\
 & Basic RD        & 0.308 \\
 & Heretic         & 0.772 \\
 & WRD-CMAES       & 0.923 \\
 & WRD-GRPO-IS     & \textbf{0.931} \\
\bottomrule
\end{tabular}
}
\caption{\new{Concept-cone comparison. Test ASR under WildGuard; WRDO rows are taken at
$\alpha = 3.0$. The WRD-GRPO-IS rows of the upper blocks repeat the
corresponding cells of Table~\ref{tab:exp1-main} and carry its three-seed
spread; the RDO-CONE and \texttt{gpt-oss-20b} rows are single runs. Higher is a stronger attack.}}
\label{tab:exp6-extra}
\end{table}

\begin{table}[!ht]
\centering
\small
\setlength{\tabcolsep}{3pt}
{
\begin{tabular}{l ccccc}
\toprule
$\beta$ & ARC-e & ARC-c & MMLU & GSM8K & PPL \\
\midrule
$1$    & 0.52 & 0.34 & 0.25 & 0.38 & 261.32 \\
$10$   & 0.51 & 0.36 & 0.34 & 0.43 & 260.13 \\
$10^2$ & 0.60 & 0.36 & 0.59 & 0.67 & 150.72 \\
$10^3$ & \textbf{0.63} & \textbf{0.38} & \textbf{0.65} & 0.83 & 11.15 \\
$10^4$ & \textbf{0.63} & \textbf{0.38} & \textbf{0.65} & \textbf{0.84} & \textbf{11.12} \\
\bottomrule
\end{tabular}
}
\caption{\new{Utility side of the $\beta$ sweep, same runs as
Table~\ref{tab:beta-asr}. Higher is better for the four accuracies, lower is
better for WikiText perplexity. Standard errors are $0.01$ for ARC-e, ARC-c and
GSM8K and $0.004$ for MMLU.
\textbf{Bold} marks the best value per column.}}
\label{tab:beta-util}
\end{table}

\end{appendixpart}

\begin{thebibliography}{30}
\providecommand{\natexlab}[1]{#1}
\providecommand{\url}[1]{\texttt{#1}}
\expandafter\ifx\csname urlstyle\endcsname\relax
  \providecommand{\doi}[1]{doi: #1}\else
  \providecommand{\doi}{doi: \begingroup \urlstyle{rm}\Url}\fi

\bibitem[Ouyang et~al.(2022)Ouyang, Wu, Jiang, Almeida, Wainwright, Mishkin,
  Zhang, Agarwal, Slama, Ray, Schulman, Hilton, Kelton, Miller, Simens, Askell,
  Welinder, Christiano, Leike, and Lowe]{ouyang2022training}
Long Ouyang, Jeff Wu, Xu~Jiang, Diogo Almeida, Carroll~L. Wainwright, Pamela
  Mishkin, Chong Zhang, Sandhini Agarwal, Katarina Slama, Alex Ray, John
  Schulman, Jacob Hilton, Fraser Kelton, Luke Miller, Maddie Simens, Amanda
  Askell, Peter Welinder, Paul Christiano, Jan Leike, and Ryan Lowe.
\newblock Training language models to follow instructions with human feedback,
  2022.

\bibitem[Bai et~al.(2022)Bai, Jones, Ndousse, Askell, Chen, DasSarma, Drain,
  Fort, Ganguli, Henighan, Joseph, Kadavath, Kernion, Conerly, El-Showk,
  Elhage, Hatfield-Dodds, Hernandez, Hume, Johnston, Kravec, Lovitt, Nanda,
  Olsson, Amodei, Brown, Clark, McCandlish, Olah, Mann, and
  Kaplan]{bai2022training}
Yuntao Bai, Andy Jones, Kamal Ndousse, Amanda Askell, Anna Chen, Nova DasSarma,
  Dawn Drain, Stanislav Fort, Deep Ganguli, Tom Henighan, Nicholas Joseph,
  Saurav Kadavath, Jackson Kernion, Tom Conerly, Sheer El-Showk, Nelson Elhage,
  Zac Hatfield-Dodds, Danny Hernandez, Tristan Hume, Scott Johnston, Shauna
  Kravec, Liane Lovitt, Neel Nanda, Catherine Olsson, Dario Amodei, Tom Brown,
  Jack Clark, Sam McCandlish, Chris Olah, Ben Mann, and Jared Kaplan.
\newblock Training a helpful and harmless assistant with reinforcement learning
  from human feedback, 2022.

\bibitem[Zou et~al.(2023)Zou, Wang, Carlini, Nasr, Kolter, and
  Fredrikson]{zou2023gcg}
Andy Zou, Zifan Wang, Nicholas Carlini, Milad Nasr, J.~Zico Kolter, and Matt
  Fredrikson.
\newblock Universal and transferable adversarial attacks on aligned language
  models, 2023.

\bibitem[Wei et~al.(2023)Wei, Haghtalab, and Steinhardt]{wei2023jailbroken}
Alexander Wei, Nika Haghtalab, and Jacob Steinhardt.
\newblock Jailbroken: How does {LLM} safety training fail?, 2023.

\bibitem[Arditi et~al.(2024)Arditi, Obeso, Syed, Paleka, Panickssery, Gurnee,
  and Nanda]{arditi2024refusal}
Andy Arditi, Oscar Obeso, Aaquib Syed, Daniel Paleka, Nina Panickssery, Wes
  Gurnee, and Neel Nanda.
\newblock Refusal in language models is mediated by a single direction, 2024.
\newblock URL \url{https://arxiv.org/abs/2406.11717}.

\bibitem[Weidmann(2025)]{heretic}
Philipp~Emanuel Weidmann.
\newblock Heretic: Fully automatic censorship removal for language models.
\newblock \url{https://github.com/p-e-w/heretic}, 2025.

\bibitem[Ahsen et~al.(2019)Ahsen, Vogel, and Stolovitzky]{ahsen2019weighted}
Mehmet~Eren Ahsen, Robert~M. Vogel, and Gustavo~A. Stolovitzky.
\newblock Unsupervised evaluation and weighted aggregation of ranked
  classification predictions.
\newblock \emph{Journal of Machine Learning Research}, 20\penalty0
  (166):\penalty0 1--40, 2019.

\bibitem[Shaikh et~al.(2026)Shaikh, Bhargavi, and Kumar~C]{shaikh2026fedwapr}
Abdullah Abdul~Sattar Shaikh, M.~S. Bhargavi, and Pavan Kumar~C.
\newblock {FedWAPR}: Bridging theory and practice in probability-driven
  weighted aggregation for federated learning.
\newblock \emph{Information Sciences}, 723:\penalty0 122697, 2026.

\bibitem[Veprikov et~al.(2026)Veprikov, Solodkin, Zyl, Savchenko, and
  Beznosikov]{veprikov2026weightlora}
Andrey Veprikov, Vladimir Solodkin, Alexander Zyl, Andrey Savchenko, and
  Aleksandr Beznosikov.
\newblock {WeightLoRA}: Keep only necessary adapters.
\newblock In \emph{Proceedings of the 64th Annual Meeting of the Association
  for Computational Linguistics (ACL)}, pages 12419--12437, 2026.

\bibitem[Wollschl{\"a}ger et~al.(2025)Wollschl{\"a}ger, Elstner, Geisler,
  Cohen-Addad, G{\"u}nnemann, and Gasteiger]{wollschlager2025geometry}
Tom Wollschl{\"a}ger, Jannes Elstner, Simon Geisler, Vincent Cohen-Addad,
  Stephan G{\"u}nnemann, and Johannes Gasteiger.
\newblock The geometry of refusal in large language models: Concept cones and
  representational independence.
\newblock \emph{arXiv preprint arXiv:2502.17420}, 2025.

\bibitem[Hansen and Ostermeier(2001)]{hansen2001completely}
Nikolaus Hansen and Andreas Ostermeier.
\newblock Completely derandomized self-adaptation in evolution strategies.
\newblock \emph{Evolutionary Computation}, 9\penalty0 (2):\penalty0 159--195,
  2001.

\bibitem[Hansen(2023)]{hansen2016cmatutorial}
Nikolaus Hansen.
\newblock The cma evolution strategy: A tutorial, 2023.
\newblock URL \url{https://arxiv.org/abs/1604.00772}.

\bibitem[Wierstra et~al.(2014)Wierstra, Schaul, Glasmachers, Sun, Peters, and
  Schmidhuber]{wierstra2014nes}
Daan Wierstra, Tom Schaul, Tobias Glasmachers, Yi~Sun, Jan Peters, and
  J{\"u}rgen Schmidhuber.
\newblock Natural evolution strategies, 2014.

\bibitem[Salimans et~al.(2017)Salimans, Ho, Chen, Sidor, and
  Sutskever]{salimans2017es}
Tim Salimans, Jonathan Ho, Xi~Chen, Szymon Sidor, and Ilya Sutskever.
\newblock Evolution strategies as a scalable alternative to reinforcement
  learning, 2017.

\bibitem[Huang et~al.(2023)Huang, Gupta, Xia, Li, and
  Chen]{huang2023catastrophic}
Yangsibo Huang, Samyak Gupta, Mengzhou Xia, Kai Li, and Danqi Chen.
\newblock Catastrophic jailbreak of open-source {LLMs} via exploiting
  generation, 2023.

\bibitem[Mazeika et~al.(2023)Mazeika, Zou, Mu, Phan, Wang, Yu, Khoja, Jiang,
  O'Gara, Xiang, Rajabi, Hendrycks, Poovendran, Li, and Forsyth]{tdc2023}
Mantas Mazeika, Andy Zou, Norman Mu, Long Phan, Zifan Wang, Chunru Yu, Adam
  Khoja, Fengqing Jiang, Aidan O'Gara, Zhen Xiang, Arezoo Rajabi, Dan
  Hendrycks, Radha Poovendran, Bo~Li, and David Forsyth.
\newblock {TDC 2023 (LLM Edition)}: The trojan detection challenge.
\newblock NeurIPS 2023 Competition Track, 2023.
\newblock URL \url{https://neurips.cc/virtual/2023/competition/66583}.

\bibitem[Mazeika et~al.(2024)Mazeika, Phan, Yin, Zou, Wang, Mu, Sakhaee, Li,
  Basart, Li, Forsyth, and Hendrycks]{mazeika2024harmbench}
Mantas Mazeika, Long Phan, Xuwang Yin, Andy Zou, Zifan Wang, Norman Mu, Elham
  Sakhaee, Nathaniel Li, Steven Basart, Bo~Li, David Forsyth, and Dan
  Hendrycks.
\newblock {HarmBench}: A standardized evaluation framework for automated red
  teaming and robust refusal, 2024.

\bibitem[Chao et~al.(2024)Chao, Debenedetti, Robey, Andriushchenko, Croce,
  Sehwag, Dobriban, Flammarion, Pappas, Tram{\`e}r, Hassani, and
  Wong]{chao2024jailbreakbench}
Patrick Chao, Edoardo Debenedetti, Alexander Robey, Maksym Andriushchenko,
  Francesco Croce, Vikash Sehwag, Edgar Dobriban, Nicolas Flammarion, George~J.
  Pappas, Florian Tram{\`e}r, Hamed Hassani, and Eric Wong.
\newblock {JailbreakBench}: An open robustness benchmark for jailbreaking large
  language models, 2024.

\bibitem[Souly et~al.(2024)Souly, Lu, Bowen, Trinh, Hsieh, Pandey, Abbeel,
  Svegliato, Emmons, Watkins, and Toyer]{souly2024strongreject}
Alexandra Souly, Qingyuan Lu, Dillon Bowen, Tu~Trinh, Elvis Hsieh, Sana Pandey,
  Pieter Abbeel, Justin Svegliato, Scott Emmons, Olivia Watkins, and Sam Toyer.
\newblock A {StrongREJECT} for empty jailbreaks, 2024.

\bibitem[Taori et~al.(2023)Taori, Gulrajani, Zhang, Dubois, Li, Guestrin,
  Liang, and Hashimoto]{taori2023alpaca}
Rohan Taori, Ishaan Gulrajani, Tianyi Zhang, Yann Dubois, Xuechen Li, Carlos
  Guestrin, Percy Liang, and Tatsunori~B. Hashimoto.
\newblock Alpaca: A strong, replicable instruction-following model, 2023.

\bibitem[Team(2025)]{qwen3technicalreport}
Qwen Team.
\newblock Qwen3 technical report, 2025.
\newblock URL \url{https://arxiv.org/abs/2505.09388}.

\bibitem[{OLMo Team}(2025)]{olmo2025olmo3}
{OLMo Team}.
\newblock Olmo 3, 2025.
\newblock arXiv:2512.13961.

\bibitem[Han et~al.(2024)Han, Rao, Ettinger, Jiang, Lin, Lambert, Choi, and
  Dziri]{han2024wildguard}
Seungju Han, Kavel Rao, Allyson Ettinger, Liwei Jiang, Bill~Yuchen Lin, Nathan
  Lambert, Yejin Choi, and Nouha Dziri.
\newblock {WildGuard}: Open one-stop moderation tools for safety risks,
  jailbreaks, and refusals of {LLMs}, 2024.

\bibitem[{Qwen Team}(2025)]{qwen3guard}
{Qwen Team}.
\newblock {Qwen3Guard} technical report, 2025.
\newblock arXiv:2510.14276.

\bibitem[Inan et~al.(2023)Inan, Upasani, Chi, Rungta, Iyer, Mao, Tontchev, Hu,
  Fuller, Testuggine, and Khabsa]{inan2023llamaguard}
Hakan Inan, Kartikeya Upasani, Jianfeng Chi, Rashi Rungta, Krithika Iyer,
  Yuning Mao, Michael Tontchev, Qing Hu, Brian Fuller, Davide Testuggine, and
  Madian Khabsa.
\newblock {LlamaGuard}: {LLM}-based input-output safeguard for human-{AI}
  conversations, 2023.

\bibitem[Gao et~al.(2024)Gao, Tow, Abbasi, Biderman, Black, DiPofi, Foster,
  Golding, Hsu, Le~Noac'h, Li, McDonell, Muennighoff, Ociepa, Phang, Reynolds,
  Schoelkopf, Skowron, Sutawika, Tang, Thite, Wang, Wang, and
  Zou]{eval-harness}
Leo Gao, Jonathan Tow, Baber Abbasi, Stella Biderman, Sid Black, Anthony
  DiPofi, Charles Foster, Laurence Golding, Jeffrey Hsu, Alain Le~Noac'h,
  Haonan Li, Kyle McDonell, Niklas Muennighoff, Chris Ociepa, Jason Phang,
  Laria Reynolds, Hailey Schoelkopf, Aviya Skowron, Lintang Sutawika, Eric
  Tang, Anish Thite, Ben Wang, Kevin Wang, and Andy Zou.
\newblock The language model evaluation harness, 07 2024.
\newblock URL \url{https://zenodo.org/records/12608602}.

\bibitem[Clark et~al.(2018)Clark, Cowhey, Etzioni, Khot, Sabharwal, Schoenick,
  and Tafjord]{clark2018arc}
Peter Clark, Isaac Cowhey, Oren Etzioni, Tushar Khot, Ashish Sabharwal, Carissa
  Schoenick, and Oyvind Tafjord.
\newblock Think you have solved question answering? try {ARC}, the {AI2}
  reasoning challenge, 2018.

\bibitem[Hendrycks et~al.(2020)Hendrycks, Burns, Basart, Zou, Mazeika, Song,
  and Steinhardt]{hendrycks2020mmlu}
Dan Hendrycks, Collin Burns, Steven Basart, Andy Zou, Mantas Mazeika, Dawn
  Song, and Jacob Steinhardt.
\newblock Measuring massive multitask language understanding, 2020.

\bibitem[{OpenAI}(2025)]{gptoss2025}
{OpenAI}.
\newblock {gpt-oss-120b} and {gpt-oss-20b} model card, 2025.

\bibitem[Akiba et~al.(2019)Akiba, Sano, Yanase, Ohta, and
  Koyama]{akiba2019optuna}
Takuya Akiba, Shotaro Sano, Toshihiko Yanase, Takeru Ohta, and Masanori Koyama.
\newblock {Optuna}: A next-generation hyperparameter optimization framework.
\newblock In \emph{KDD}, 2019.

\end{thebibliography}
\end{document}